\documentclass[sigconf]{acmart}

\usepackage{threeparttable}
\usepackage{amsmath}
\usepackage{algorithm}
\usepackage{algpseudocode}
\usepackage{array}
\usepackage{booktabs}
\usepackage{multirow}
\usepackage{enumitem}
\usepackage{float}
\usepackage{multicol,lipsum}
\usepackage{graphicx}
\usepackage{placeins}
\usepackage{hyperref}
\usepackage[commandnameprefix=always]{changes}
\usepackage{todonotes}
\usepackage{balance}
\usepackage{soul}

\AtBeginDocument{%
  \providecommand\BibTeX{{%
    \normalfont B\kern-0.5em{\scshape i\kern-0.25em b}\kern-0.8em\TeX}}}

\copyrightyear{2024}
\acmYear{2024}
\setcopyright{rightsretained}
\acmConference[CIKM '24]{Proceedings of the 33rd ACM International Conference on Information and Knowledge Management}{October 21--25, 2024}{Boise, ID, USA}
\acmBooktitle{Proceedings of the 33rd ACM International Conference on Information and Knowledge Management (CIKM '24), October 21--25, 2024, Boise, ID, USA}
\acmDOI{10.1145/3627673.3679844}
\acmISBN{979-8-4007-0436-9/24/10}

\begin{document}

\title{Time is Not Enough: Time-Frequency based Explanation for Time-Series Black-Box Models}

\author{Hyunseung Chung}
\affiliation{
  \institution{KAIST}
  \city{Daejeon}
  \country{Republic of Korea}
}
\email{hs_chung@kaist.ac.kr}

\author{Sumin Jo}
\affiliation{
  \institution{KAIST}
  \city{Daejeon}
  \country{Republic of Korea}
}
\email{ekrxjwh2009@kaist.ac.kr}

\author{Yeonsu Kwon}
\affiliation{
  \institution{KAIST}
  \city{Daejeon}
  \country{Republic of Korea}
}
\email{yeonsu.k@kaist.ac.kr}

\author{Edward Choi}
\affiliation{
  \institution{KAIST}
  \city{Daejeon}
  \country{Republic of Korea}
}
\email{edwardchoi@kaist.ac.kr}
                                                                  
\renewcommand{\shortauthors}{Hyunseung Chung, Sumin Jo, Yeonsu Kwon, and Edward Choi}

\begin{abstract}
Despite the massive attention given to time-series explanations due to their extensive applications, a notable limitation in existing approaches is their primary reliance on the time-domain.
This overlooks the inherent characteristic of time-series data containing both time and frequency features.
In this work, we present Spectral eXplanation (SpectralX), an XAI framework that provides time-frequency explanations for time-series black-box classifiers.
This easily adaptable framework enables users to ``plug-in'' various perturbation-based XAI methods\footnote{In this work, we distinguish between XAI \textit{framework} and XAI \textit{method}.} for any pre-trained time-series classification models to assess their impact on the explanation quality without having to modify the framework architecture.
Additionally, we introduce Feature Importance Approximations (FIA), a new perturbation-based XAI method.
These methods consist of feature \textit{insertion, deletion}, and \textit{combination} techniques to enhance computational efficiency and class-specific explanations in time-series classification tasks. 
We conduct extensive experiments in the generated synthetic dataset and various UCR Time-Series datasets to first compare the explanation performance of FIA and other existing perturbation-based XAI methods in both time-domain and time-frequency domain, and then show the superiority of our FIA in the time-frequency domain with the SpectralX framework. 
Finally, we conduct a user study to confirm the practicality of our FIA in SpectralX framework for class-specific time-frequency based time-series explanations. The source code is available  \href{https://github.com/gustmd0121/Time_is_not_Enough}{\textbf{here}}.        
\end{abstract}



\maketitle

\begin{figure}[H]
  \centering
  \includegraphics[width=\linewidth]{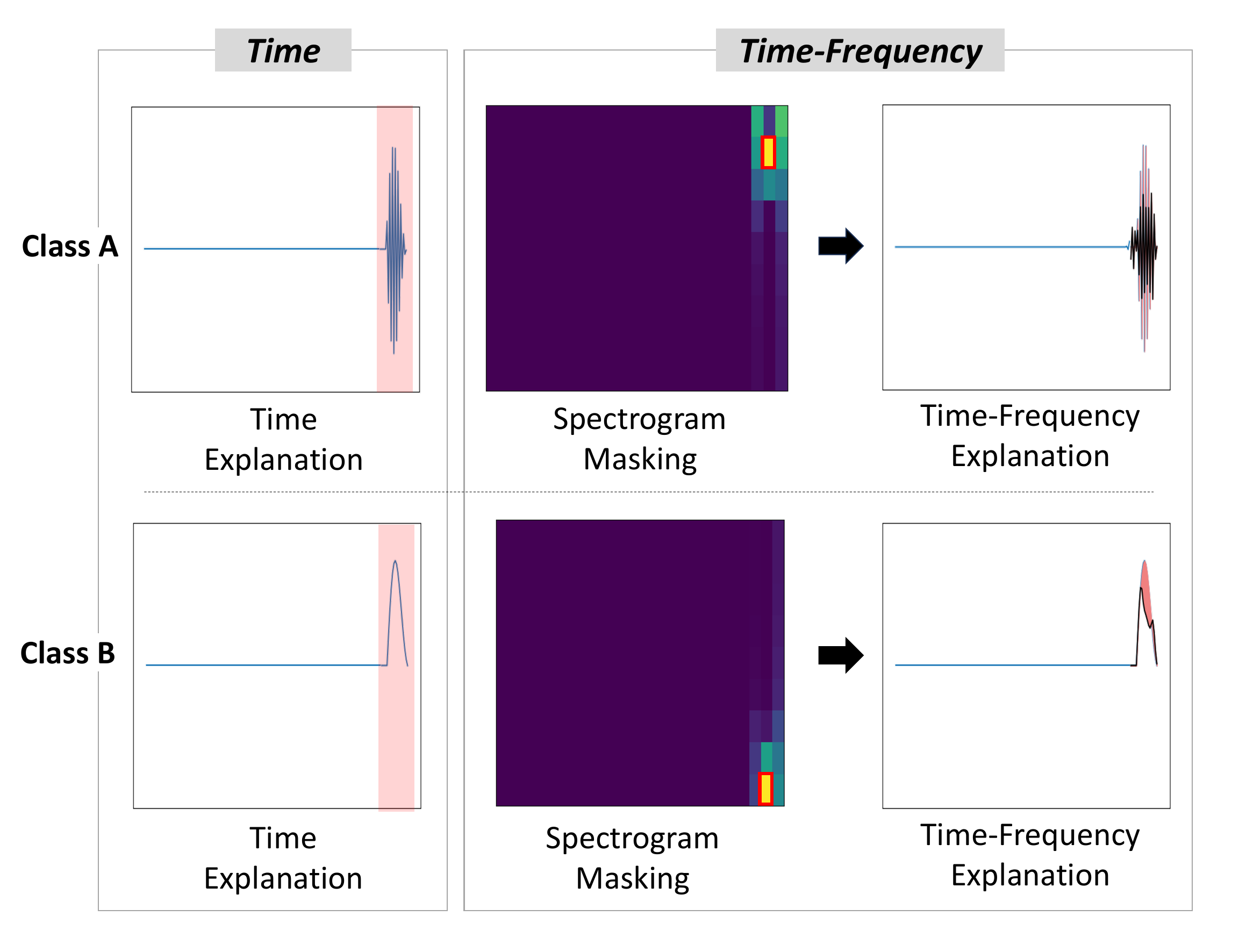}
  \caption{Example of the time-domain explanation, where the important regions are overlapping for both classes. In time-frequency domain explanation, however, zero-masking the red-box, which indicates important region in the time-frequency domain, appears as the red region in original time-domain which explains the most important feature of the class.}
  \Description{Example of time-domain explanation, where the important regions overlap for both classes. In the time-frequency domain explanation, however, the important regions do not overlap and the impact of the frequency component is shown in the time-domain by masking the region.}
  \label{motivational}
\end{figure}

\section{Introduction}

In recent years, the field of Time-Series Classification (TSC) grew rapidly due to the widespread availability of open-source time-series datasets across various domains, and significant advancements in deep learning classification models. These models, particularly Long Short-Term Memory (LSTM) networks \cite{hochreiter1997long}, Convolutional Neural networks (CNNs) \cite{lecun2015deep}, and Transformers \cite{vaswani2017attention}, have been crucial in effectively capturing intricate temporal features in time-series data. As the complexity of these models increased, their classification accuracy also drastically improved. However, this rapid escalation in complexity overall had a detrimental impact on model interpretability, making these ``black box'' models less transparent and harder to understand \cite{adadi2018peeking}.

In light of these complexities, the need for Explainable AI (XAI) has become vital in these models because of the high-stake nature of decisions made in many critical domains of TSC such as healthcare, finance, and climate science.
Especially in these cases, explainability of classification models is necessary to understand and validate the rationale behind a model's decision.
Previous methods for XAI on time-series black-box classifiers are categorized into two categories: 1) Adopting representative perturbation-based XAI methods from other domains (image, text, tabular) to the time-series domain \cite{schlegel2019towards, jalali2020machine, schlegel2023deep}; and 2) Creating novel time-series specific perturbation techniques \cite{8995349, neves2021interpretable, sivill2022limesegment, raab2023xai4eeg, mishra2020reliable}. 

To expand on the first category, there were many representative perturbation-based XAI methods in other domains.
However, these methods were not originally designed for the time-series domain, and therefore were only applicable to domains such as tabular, image, or text.
In order to utilize these methods for time-series model explanations, various adaptation efforts and rigorous evaluations were required because the nature of time-series data was unlike any other data types.
The common feature among the previous works that successfully adopted popular methods from other domains to time-series \cite{schlegel2019towards, jalali2020machine, schlegel2023deep} is that they were primarily model-agnostic perturbation-based XAI methods such as RISE \cite{petsiuk2018rise}, LIME \cite{ribeiro2016should}, or SHAP \cite{lundberg2017unified}.
Although these methods captured important features in the time-domain and generated heatmaps according to feature importance weights, they did not consider the frequency component of time-series data. 
       
After this phase of adoption, researchers began to introduce various time-series specific perturbation-based XAI methods. These methods focused on considering the distinct characteristics of time-series data such as temporal dependence \cite{8995349, neves2021interpretable, sivill2022limesegment}. 
For time-series classifier models, these methods provided better explanation performance compared to previous methods. However, explanations were limited to determining important regions in the time-domain based on temporal patterns and trends.
They did not consider frequency components of signals, which contained rich information on transient events or periodic patterns that were potentially influential to the model's decision-making process.  

Time-frequency analysis provides a more comprehensive view of the time-series data by capturing both temporal and spectral features.
Most real-world time-series datasets contain time-varying spectral properties, which implies the frequency components change over time.
In other words, utilizing time-frequency features for model decision explanation generates better time-series explanation than solely relying on time-domain features.
An example is shown in Figure \ref{motivational}, which depicts a binary class time-series dataset consisting of same important time-regions but different frequency components for each class.
The time-domain explanation highlights the same regions in class A and class B as important, which does not provide valuable class-specific explanation.
On the other hand, the time-frequency explanation shows the impact of the most important frequency component for each class by visualizing the difference between with and without the most important frequency component in the time-domain signal.
Similar to this example, there are many cases where the frequency band is a crucial feature in distinguishing between classes for a time-series classification model. For example in sleep EEG dataset \cite{kemp2000analysis}, the sleep stage classes are distinguished by frequency ranges, and in ECG dataset \cite{moody1983new}, Atrial fibrillation contains different frequency range compared to normal.                      

In this paper we propose Spectral eXplanation, an XAI framework to provide time-frequency explanations of time-series black-box classifier models. SpectralX enables any perturbation-based XAI method (e.g. LIME, KernelSHAP, RISE) or black-box classifier models (e.g. LSTM, CNN, Transformer) to ``plug-and-play'' in our framework to determine important time-frequency explanations.
Through our framework we are able to determine the critical time-series regions by observing the frequency components over the change in time.
Given a time-series classification dataset, SpectralX performs a Short-Time Fourier Transform (STFT), search for important class-wise features in the time-frequency domain via feature perturbation, then perform an inverse operation to revert back to the original raw time-series domain to determine class probabilities of class-specific perturbations.
Ultimately, SpectralX maintains the black-box classifier model input to be raw time-series signals, and output to represent probability logits of each class of the time-series dataset.

In addition to SpectralX, we propose a set of new perturbation-based XAI methods named Feature Importance Approximations (FIA).
The FIA consist of \textit{insertion}, \textit{deletion}, and \textit{combined} methods.
The main objective of these methods is to approximate the importance of features in the black-box model's decision-making process by quantifying the significance of features in contributing to the model's prediction.
In insertion, this is approximated by progressively introducing significant features into a baseline version of the data, and observing the impact on the model's output.
The deletion method progressively removes features from the original data, and highlights features whose absence most negatively affect the model predictions.
The combined method takes the absolute difference between both insertion and deletion methods, considering both positive impact when inserted and the negative impact when deleted to determine important features.    

Our contributions can be summarized as follows: 
\begin{itemize}[leftmargin=3mm]
    \item To the best of our knowledge we propose the first general XAI framework, SpectralX, leveraging both time and frequency information to explain black-box time-series classifiers for various time-series datasets.
    Our framework facilitates straightforward integration of diverse perturbation-based XAI methods and classifiers to assess spectral explanation performance.     
    
    \item We introduce FIA, which surpasses existing sample-wise perturbation-based XAI methods in efficiency and effectiveness. The FIA is designed to be class-specific, thus enhancing computational performance with direct approximations of important features in each class.

    \item We perform extensive experiments across a vast array of time-series datasets to empirically validate the effectiveness of SpectralX and the FIA. These experiments demonstrate significant improvements enabled by our work over existing methods in terms of time-series explanation.

    \item We conduct a user study to evaluate the practical applicability and user perception of FIA.
    The results of this study affirm the practicality of our proposed perturbation method, highlighting their potential impact in real-world applications.
\end{itemize}

\begin{figure*}[t]
  \centering
  \includegraphics[width=\textwidth]{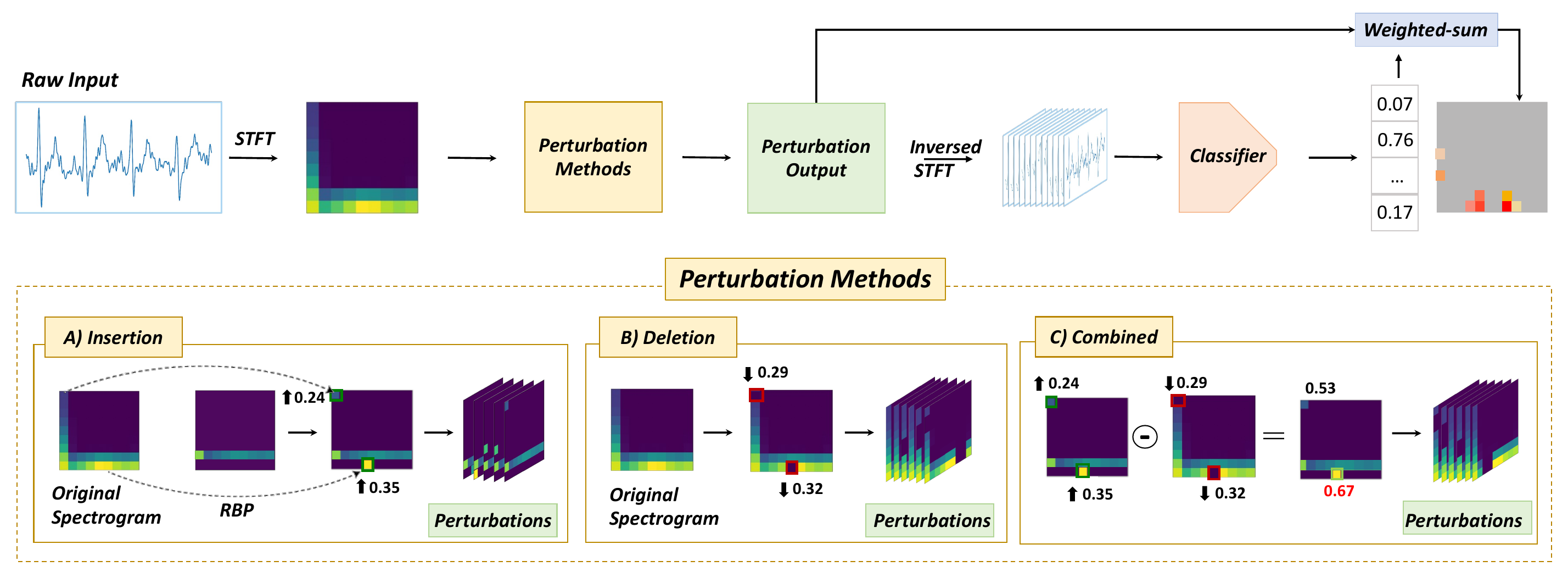}
  \caption{(\textit{Top}) SpectralX Framework and Feature Importance Approximations (FIA). The raw signals are converted to time-frequency representations with Short-Time Fourier Transform(STFT), and a single perturbation-based XAI method is used to generate perturbation outputs. After reverting to perturbed time signal with Inverse STFT, the classifier determines probability scores to determine important class-specific time-frequency features. 
  (\textit{Bottom}) The three perturbation-based XAI methods represent our FIA. The Insertion method introduces significant features into RBP (Realistic Background Perturbation), a global property of the signal, the Deletion method removes significant features from the original signal, and the Combined method merges the two methods. Numeric values represent increase or decrease in class probability for each method.}
  \Description{The overall architecture of our proposed Spectral eXplanation framework. For the Feature Importance Approximations (FIA), visualizations of the Insertion, Deletion, and Combined methods.}
  \label{framework}
\end{figure*}

\section{Related Work}
\label{sec:relatedwork}
XAI methods for time-series models can generally be divided into three different categories: Backpropagation-based methods, attention-based methods, and perturbation-based methods.  

For the first two categories, there were many notable studies such as Layer-wise Relevance Propagation (LRP) \cite{bach2015pixel}, which decomposed a neural network's output through the layers, Integrated Gradients \cite{sundararajan2017axiomatic}, which quantified importance of model's prediction by integrating the gradients, and Grad-CAM \cite{selvaraju2017grad} that produced heatmap visualizations with gradients for CNNs. For time-series models, Dynamic Masking \cite{crabbe2021explaining} used backpropagation to selectively focus on relevant portions of the data, allowing models to adaptively emphasize important temporal patterns. Also, attention was used to interpret a CNN-LSTM hybrid TSC model in \cite{karim2017lstm}, and temporal contextual layers were introduced for explanation in TSC \cite{vinayavekhin2018focusing, hsieh2021explainable}.   
However all these studies require a specific model architecture, or bear a clear constraint (\textit{e.g.} must be a CNN-based model, cannot support decision tree-based models, etc.).
In this work, we aim to develop a model-agnostic XAI framework for time-series classifiers, which naturally sets our focus on perturbation-based methods. Therefore, for the rest of this work, we mainly deal with perturbation-based methods.

For perturbation-based methods, representative methods in the general domain were adapted to the time-series domain.
There were numerous studies of perturbation-based methods in the general domain \cite{duell2021comparison, ribeiro2016should, gramegna2021shap, makridis2023xai, lundberg2017unified, saluja2021towards, petsiuk2018rise}. Among these approaches, methods like LIME \cite{ribeiro2016should}, SHAP \cite{lundberg2017unified}, and RISE \cite{petsiuk2018rise} have been prominently acknowledged for their contributions to the understanding of model explainability. 

LIME \cite{ribeiro2016should} presents a model-agnostic black-box strategy by creating random samples around an instance to be explained, and construct simpler, linear decision models. This approach aids in clarifying complex models through local interpretability.
SHAP \cite{lundberg2017unified} merges concepts from cooperative game theory with local explanations. It assigns the prediction output to the contributions of each feature for a given instance, offering a detailed breakdown of feature importance in model predictions.
RISE \cite{petsiuk2018rise} is particularly useful in interpreting predictions of black-box models in visual data. It works by randomly masking parts of the input image and observing the changes in the model's output. This results in a saliency map that highlights key regions in the input that significantly influence the model's decisions. 

While these methodologies have provided valuable insights, directly applying them to time series classifiers neglected the unique aspects of time series data, notably their temporal dependence. This risked misguiding interpretations of model's decisions. Recognizing this gap, the focus has recently shifted towards developing perturbation methods that are specifically tailored for time series classifiers. 
For instance, the study by Neves et al. \cite{neves2021interpretable} proposed enhancing time series explainability by integrating raw signals with their derivatives. Also, the authors in LIMESegment \cite{sivill2022limesegment} adapt LIME for time series analysis, utilizing NNSegment, harmonic perturbation, and Dynamic Time Warping. LEFTIST \cite{8995349} also presented local time series classifiers, focusing on a systematic approach that includes segmenting components, generating neighbors.

Although most existing research in this field focuses solely on the time-domain, it's equally important to take into account both time and frequency. The rationale behind this is that frequency, just like time, is a fundamental and crucial property inherent in time series data \cite{nguyen2018interpretable}, \cite{wang2023wavelet}, \cite{wu2023time}.
Also, most real-world time-series datasets contain dynamic frequency behaviors, which can  easily be tracked in the time-frequency domain. This is because dynamic frequency behaviors refer to frequency change over time, which is more easily identifiable in time-frequency domain with observation of transient behaviors and periodic patterns. Therefore, studies have emerged that focus on time-frequency, but the research by Raab et al.\cite{raab2023xai4eeg} is limited to EEG data and has the drawback of requiring separate black-box classifiers trained for temporal and spatio-temporal domain explanation. Moreover, Mishra et al. \cite{mishra2020reliable} focused their research solely on audio data, which hinders the model's adaptability to a wider array of general time-series dataset explanations.

Our main objective of the SpectralX framework is to create a versatile XAI framework that can integrate any TSC model for time-frequency explanations in general time-series datasets.
Therefore, as mentioned above, we utilize various perturbation-based methods in our framework to experimentally compare time-frequency explanations in diverse time-series datasets. Furthermore, we present a novel FIA that improves upon other baseline perturbation-based methods.  

\section{SpectralX}
The SpectralX framework is designed to provide spectral explanations for time-series classifiers trained on raw signal data.
The highly adaptable characteristic of this framework enables users to ``plug-in'' various perturbation-based XAI methods and pre-trained TSC models for time-frequency explanations.
The extensive performance evaluations across diverse perturbation methods\footnote{We use interchangeably, perturbation methods, perturbation-based methods, and perturbation-based XAI methods.} and classifier models ensure results are robust and generalizable.
The overall framework is shown in Figure \ref{framework}. In the following sections we will explore the step-by-step process of SpectralX in detail.     

\subsection{Short-Time Fourier Transform}
The dynamic frequency behavior of real-world time-series datasets diversify the frequency component within each time-series data. Therefore, analyzing the change in frequency over time is crucial in determining class-specific explainable characteristics in time-series datasets. Fortunately, the Short-Time Fourier Transform (STFT) captures the time-varying frequency content of time-series signals. In SpectralX, the STFT is employed as a preliminary step before applying a perturbation method. The equation for STFT is given as:

\begin{equation}
S[m,k] = \sum_{n=0}^{N-1} x[n+mH] \cdot w[n] \cdot e^{-j2\pi \frac{k}{N}n}
\end{equation}

where \(S\) is the total STFT output and \(S \in \mathbb{C}^{M \times K}\), $M$ representing the number of total time-segments and $K$ representing the total number of frequency bins. 
Also, $N$ is the number of samples in the window, $w[n]$ denotes the window function applied to $n$, which is within the windowed portion of the signal. The term $x[n+mH]$ indicates that for each time segment $m$, multiplied by the hop size $H$, the window is applied to a different segment of the signal. The exponential term $e^{-j2\pi nk/N}$ is the Fourier Transform kernel for a discrete frequency $k$. Essentially, the STFT output is the frequency content of the signal over time.              

\subsection{Perturbation-based XAI Methods}
As mentioned in the \hyperref[sec:relatedwork]{Related Work} section, our primary objective of creating a framework that supports any TSC model necessitates the use of perturbation methods out of the three different categories of time-series explanation methods.
Therefore, SpectralX framework is specifically designed to integrate any perturbation method with any TSC model. 

Following the transformation of raw time-series data into time-frequency representations via STFT, SpectralX employs a perturbation method to modify time-frequency components of them.
The main goal of this process is to assess the impact of these perturbations on particular regions of the STFT output (\textit{i.e.} time-frequency representation), identifying the most influential areas.
In random regions in the STFT output, a predetermined number of different perturbations are applied.
Afterwards, these perturbations are reverted back to the time-domain to determine the class probability in the pre-trained black-box classifier.
The initial probability of the original data is compared to the perturbed probabilities to determine salient features.    

\subsection{Inverse Short-Time Fourier Transform}
The reverse process from time-frequency domain to time domain is accomplished through the inverse Short-Time Fourier Transform (ISTFT).
During this process, it is crucial to ensure perfect reconstruction. Otherwise, noise generated during conversion can interfere with explanations.
There are three conditions that must be met in order to ensure perfect reconstruction.
First, the Overlap-Add (OLA) condition requires 50 percent overlap between window segments for the Hann window \cite{harris1978use}, which is the smoothing window function used for both STFT and ISTFT.
Second, the window function used in the STFT must match the window function used in ISTFT.
Finally, both magnitude and phase information must be correctly incorporated as input to the ISTFT.
With all these conditions met, our ISTFT produces time-domain signals with no information loss or distortion, except the alterations from time-frequency perturbations.
The equation for ISTFT is as follows: 

\begin{equation}
\hat{x}[n] = \frac{1}{N}\sum_{m=0}^{M-1}\sum_{k=0}^{N-1} S[m,k] \cdot w[n-mH] \cdot e^{+j2\pi \frac{k}{N}n}
\end{equation}

Here, \(\hat{x} \in \mathbb{R}^{L}\), where $L$ represents the total one-dimensional length of the reconstructed time-series signal. $M$ is the total number of windows in the STFT, and all other variables are the same as in STFT. After the ISTFT process, perturbations reverted to time-domain signals are used as inputs to the black-box TSC classifier, to determine important time-frequency features in various ways depending on the perturbation methods. 

\section{Feature Importance Approximation}
The impact of feature insertion and deletion towards the change in class probability served as a key indicator of the quality of explanations \cite{ivanovs2021perturbation}.
Several evaluation metrics for explanation were proposed based on these probability variations \cite{fong2017interpretable, petsiuk2018rise, theissler2022explainable}.
We propose Feature Importance Approximations (FIA) based on this concept of feature insertion and deletion, where steep rise in class probability after insertion and sharp drop in class probability after deletion indicate important class-specific features and good explanation. 
Additionally, we propose a combination between insertion and deletion methods, where the weighted absolute difference between insertion and deletion probability changes are selected.
The FIA is conducted for all samples in the test set that is in the target explanation class, and is visualized in Figure \ref{framework}.
We will describe in-detail the insertion, deletion, and combined methods. 

{\small
\renewcommand{\arraystretch}{0.9}
\begin{table}
  \caption{Average Faithfulness comparison of various methods in Time and Time-Frequency Domain for all nine UCR datasets. Mean and standard deviation are shown across three classifiers. The \textbf{boldface} highlights the best performance in each domain.}
  \setlength{\tabcolsep}{10pt}
  \label{tab:faithfulness_comparison}
  \begin{tabular}{ccc}
    \toprule
    \multicolumn{3}{c}{Average Faithfulness} \\
    \hline\hline 
    Methods & Time & Time-Frequency \\

    \hline
  \textit{Baselines}
  \\\hline
    LIME & $0.139 \pm 0.05$ & $0.146 \pm 0.06$ \\
    KernelSHAP & $0.131 \pm 0.02$ & $0.128 \pm 0.06$ \\
    RISE & $0.135 \pm 0.04$ & $0.143 \pm 0.05$ \\
    LIMESegment & $\bf0.148 \pm 0.01$ &$-$ \\
  \hline
  \noalign{\smallskip} 
  \textit{FIA (ours)}
  \\\hline
    Insertion & $0.107 \pm 0.07$ & $0.115 \pm 0.08$ \\
    Deletion & $\bf0.148 \pm 0.04$ & $0.149 \pm 0.08$ \\
    Combined & $ 0.147 \pm 0.04$ & $\bf 0.154 \pm 0.08$ \\
    \hline
    \bottomrule
  \end{tabular}
\end{table}
}

{\small
\renewcommand{\arraystretch}{0.9}
\begin{table}
  \caption{Average Robustness comparison of various methods for top-8 features in Time and Time-Frequency Domain for all nine UCR datasets. Mean and standard deviation are shown across three classifiers. The \textbf{boldface} highlights the best performance in each domain.}
  \setlength{\tabcolsep}{10pt}
  \label{tab:robustness}
  \begin{tabular}{ccc}
    \toprule
    \multicolumn{3}{c}{Average Robustness} \\
    \hline\hline 
    Methods & Time & Time-Frequency \\

    \hline
  \textit{Baselines}
  \\\hline
    LIME & $0.58 \pm 0.12$ & $0.69 \pm 0.14$ \\
    KernelSHAP & $0.59 \pm 0.13$ & $ 0.70 \pm 0.13$ \\
    RISE & $0.56 \pm 0.11$ & $0.68 \pm 0.16$ \\
    LIMESegment & $\bf0.60 \pm 0.15$ &$-$ \\
  \hline
  \noalign{\smallskip} 
  \textit{FIA (ours)}
  \\\hline
    Insertion & $0.56 \pm 0.18$ & $0.70 \pm 0.14$ \\
    Deletion & $0.58 \pm 0.16$ & $0.69 \pm 0.18$ \\
    Combined & $0.57 \pm 0.11$ & $\bf0.72 \pm 0.12$ \\
    \hline
    \bottomrule
  \end{tabular}
\end{table}
}
\subsection{Insertion}
The insertion method is designed to determine the significance of time-frequency features in enhancing the prediction accuracy of a target class by iteratively inserting features.
This method begins with the initial state where the original signal is converted to the baseline representation, which is Realistic Background Perturbation (RBP) \cite{sivill2022limesegment}.
The RBP represents a global property of the signal in the time-frequency representation, and is generated by identifying the frequency band with highest representation and lowest variance.
The dimension of the RBP output is the same as the STFT output, which is \({M \times K}\). This reflects the true background content of time-series signals, enforcing meaningful perturbations.
We denote the function as $RBP(\cdot)$, which is the RBP function to generate baseline representations.

During the insertion phase, we generate diverse perturbations by selecting random features. The selected features remain consistent across all samples within the batch to apply batch parallel processing.
This allows for more efficient computation compared to other methods such as LIME \cite{ribeiro2016should}, SHAP \cite{lundberg2017unified}, or RISE \cite{petsiuk2018rise} that either perturb or select different set of random features for each sample.
After applying ISTFT, the probability of the target class is calculated using the black-box classifier \(C\). The notation and steps for insertion are as follows:

$F$ is the set of all features (i.e., entries) in $S$, and $F_{\text{selected}}$ is the set of features selected.
Initially, we apply the $RBP$ function to the original signal $x$ to obtain $RBP(x)$ which is \(S_{\text{RBP}}\). Then, for each iteration \(i\) where \(i = 1, 2, \ldots, F\), we generate \(P\) different random binary perturbation masks, \(U^{(p)}\), for \(p = 1, 2, \ldots, P\). Each binary perturbation mask has \(R\) time-frequency regions unmasked (set to 1) while the rest are masked (set to 0). We apply each binary perturbation mask to the RBP representation to create perturbed versions, \(S_{\text{ins-per}}^{(i, p)}\), for each perturbation \(p\) at iteration \(i\). The formula is as follows:

\begin{equation}
S_{\text{ins-per}}^{(i, p)} = S \odot U^{(i,p)} + S_{\text{RBP}} \odot (1 - U^{(i,p)})
\end{equation}

Afterwards, we convert each perturbation back to the time domain using ISTFT to get \(\hat{x}_{\text{ins-per}}^{(i, p)}\). We pass each perturbed signal through the black-box classifier \(C\) to obtain class probability scores. The target class probability score of the original signal is denoted as \(P_{\text{class}}^{\text{original}}\), and the target class probability score of the perturbed signal is \(C(\hat{x}_{\text{ins-per}}^{(i, p)})\) which is represented as \(P_{\text{class}}^{\text{perturbed}}\). 
The features that are not already selected, represented by $f \in F \setminus F_{\text{selected}}$, are the features considered for importance scoring in iteration $i$. We calculate the final score based on the influence on the target class probability scores, which is: 

\begin{equation}
\text{Score}_{ins-f}^{(i)} = \frac{\sum_{p=1}^{P} \delta(U^{(i,p)}, f) \cdot (P_{\text{class}}^{\text{perturbed},f} - P_{\text{class}}^{\text{original}})}{\sum_{p=1}^{P} \delta(U^{(i,p)}, f)}
\end{equation}

\noindent where \(\delta(U^{(i,p)}, f)\) is an indicator function that equals 1 if the feature is unmasked in the \(p\)-th perturbation, and 0 otherwise. Finally, we select the feature with the highest score:
\begin{equation}
f_{\text{max}}^{(i)} = \arg\max_{f \in F \setminus F_{\text{selected}}} \text{Score}_{ins-f}^{(i)}
\end{equation}

\noindent and update $F_{\text{selected}}$ and $F$ for the next iteration as follows: 
\begin{equation}
F_{\text{selected}}^{(i+1)} = F_{\text{selected}}^{(i)} \cup \{f_{\text{max}}^{(i)}\}, \quad F^{(i+1)} = F^{(i)} \setminus \{f_{\text{max}}^{(i)}\}
\end{equation}

{\small
\renewcommand{\arraystretch}{0.95}
\begin{table*}[ht]
  \caption{Rank-Biased Overlap@k and Area Under Curves comparison of various methods in the Time-Frequency Domain for the Synthetic Dataset. Mean and standard deviation are shown across three classifiers. The \textbf{boldface} highlights the best performance in each @k, Area Under Precision (AUP), and Area Under Recall (AUR).}\label{tab:rbo}
  \setlength{\tabcolsep}{6pt}
  \begin{tabular}{lccccccc}
  \toprule
    & \multicolumn{5}{c}{Rank-Biased Overlap@k} & \multicolumn{2}{c}{Area Under Curves} \\
    \cmidrule(lr){2-6} \cmidrule(lr){7-8}
    & @1 & @2 & @4 & @6 & @8 & AUP & AUR \\
  \midrule
  \hline
  \textit{Baselines} & & & & & & & \\
  \hline
  LIME & $0.311 \pm 0.12$ & $0.311 \pm 0.13$ & $0.471 \pm 0.10$ & $0.515 \pm 0.10$ & $0.545 \pm 0.11$ & $0.489 \pm 0.05$ & $0.431 \pm 0.06$ \\
  \hline
  KernelSHAP & $\bf0.422 \pm 0.19$ & $0.367 \pm 0.14$ & $0.445 \pm 0.02$ & $0.465 \pm 0.04$ & $0.466 \pm 0.08$ & $0.444 \pm 0.06$ & $0.433 \pm 0.07$ \\
  \hline
  RISE & $0.2 \pm 0.03$ & $0.228 \pm 0.05$ & $0.357 \pm 0.13$ & $0.416 \pm 0.12$ & $0.447 \pm 0.11$ & $0.391 \pm 0.11$ & $0.330 \pm 0.07$ \\
  \midrule
  \textit{FIA (Ours)} & & & & & & & \\
  \hline
  Insertion & $0.225 \pm 0.05$ & $0.083 \pm 0.08$ & $0.206 \pm 0.12$ & $0.318 \pm 0.13$ & $0.407 \pm 0.12$ & $0.293 \pm 0.08$ & $0.248 \pm 0.06$ \\
  \hline
  Deletion & $0.333 \pm 0.07$ & $0.406 \pm 0.05$ & $0.433 \pm 0.05$ & $0.474 \pm 0.11$ & $0.491 \pm 0.09$ & $0.467 \pm 0.07$ & $0.427 \pm 0.12$ \\
  \hline
  Combined & $0.333 \pm 0.03$ & $\bf0.444 \pm 0.12$ & $\bf0.527 \pm 0.12$ & $\bf0.574 \pm 0.12$ & $\bf0.580 \pm 0.08$ & $\bf0.553 \pm 0.10$ & $\bf0.491 \pm 0.04$ \\
  \hline
  \bottomrule
  \end{tabular}
\end{table*}
}
\subsection{Deletion}
The deletion method focuses on assessing the impact of removing time-frequency features. It emphasizes the importance of features whose absence significantly degrades model performance. Instead of starting with RBPs, the deletion method starts with time-frequency representations of original signals. Also, instead of unmasking specific features, the deletion method masks \(R\) different regions to create \(P\) different perturbations. Apart from this, all other notations and steps are similar to the insertion method. The steps are as follows:   

We start with the STFT output of the signal, denoted as \(S\). For each iteration \(i\) (where \(i = 1, 2, \ldots, F\),  we generate \(P\) different random masked features, \(M^{(p)}\), for \(p = 1, 2, \ldots, P\). Each perturbation has \(R\) time-frequency regions masked (set to 0) while the rest are unmasked (set to 1). Then, for each mask \(p\) at iteration \(i\):

\begin{equation}
S_{\text{del-per}}^{(i, p)} = S \odot (1 - M^{(i,p)}) + S_{\text{RBP}} \odot M^{(i,p)}
\end{equation}

The time-frequency perturbation is reverted back to the time-domain signal \(x_{\text{del-per}}^{(i, p)}\) with ISTFT. We pass each perturbed signal through the black-box classifier to obtain target class probability scores, \(P_{\text{class}}^{(i, p)}\), for each perturbation. The change in probability scores, compared to the original signal's target class probability score, \(P_{\text{class}}^{\text{original}}\), indicates the importance of the masked regions. The final score for each feature masked in iteration \(i\) is calculated as follows:

\begin{equation}
\text{Score}_{del-f}^{(i)} = \frac{\sum_{p=1}^{P} \delta(M^{(i,p)}, f) \cdot (P_{\text{class}}^{\text{perturbed},f} - P_{\text{class}}^{\text{original}})}{\sum_{p=1}^{P} \delta(M^{(i,p)}, f)}
\end{equation}

\noindent where \(\delta(M^{(i,p)}, f)\) is an indicator function that equals 1 if the feature is unmasked, and 0 otherwise in the \(p\)-th perturbation. We select the feature with the lowest score:
\begin{equation}
f_{\text{min}}^{(i)} = \arg\min_{f \in F \setminus F_{\text{selected}}} \text{Score}_{del-f}^{(i)}
\end{equation}

\noindent and update $F_{\text{selected}}$ and $F$ for the next iteration as follows: 
\begin{equation}
F_{\text{selected}}^{(i+1)} = F_{\text{selected}}^{(i)} \cup \{f_{\text{min}}^{(i)}\}, \quad F^{(i+1)} = F^{(i)} \setminus \{f_{\text{min}}^{(i)}\}
\end{equation}

{\small
\renewcommand{\arraystretch}{0.95}
\begin{table*}[ht]
  \caption{Faithfulness@k comparison of various methods in the Time-Frequency Domain for nine UCR datasets. Mean and standard deviation are shown across three classifiers. The \textbf{boldface} highlights the best performance in each @k.}\label{tab:faithfulness@k}
  \setlength{\tabcolsep}{10pt}
  \begin{tabular}{lccccc}
  \toprule
    & \multicolumn{5}{c}{Faithfulness@k} \\
    \cmidrule(lr){2-6}
    & @1 & @2 & @4 & @6 & @8 \\
  \midrule
  \hline
  \textit{Baselines} & & & & & \\
  \hline
  LIME & $0.146 \pm 0.06$ & $0.196 \pm 0.08$ & $0.245 \pm 0.06$ & $0.285 \pm 0.09$ & $0.312 \pm 0.07$ \\
  \hline
  KernelSHAP & $0.128 \pm 0.06$ & $0.147 \pm 0.09$ & $0.204 \pm 0.08$ & $0.212 \pm 0.10$ & $0.278 \pm 0.08$ \\
  \hline
  RISE & $0.128 \pm 0.06$ & $\bf0.202 \pm 0.08$ & $0.258 \pm 0.09$ & $0.287 \pm 0.09$ & $0.332 \pm 0.07$ \\
  \midrule
  \textit{FIA (Ours)} & & & & & \\
  \hline
  Insertion & $0.115 \pm 0.08$ & $0.154 \pm 0.09$ & $0.199 \pm 0.09$ & $0.229 \pm 0.07$ & $0.273 \pm 0.04$ \\
  \hline
  Deletion & $0.149 \pm 0.08$ & $0.189 \pm 0.05$ & $0.238 \pm 0.06$ & $0.292 \pm 0.05$ & $0.333 \pm 0.08$ \\
  \hline
  Combined & $\bf0.154 \pm 0.08$ & $0.192 \pm 0.06$ & $\bf0.263 \pm 0.06$ & $\bf0.295 \pm 0.08$ & $\bf 0.343 \pm 0.06$ \\
  \hline
  \bottomrule
  \end{tabular}
\end{table*}
}

\subsection{Combined}
The combined method integrates both insertion and deletion methods. It first runs insertion and deletion method in parallel, going through the ISTFT, perturbation process, and importance scoring calculation in both methods. 
Then, to determine the final score in the combined method for each iteration, the weighted absolute difference between the final scores from insertion and deletion methods are calculated.
The time-frequency feature that produces the greatest weighted absolute difference score is the selected feature for the iteration $i$.
The formula is as follows: 

\begin{equation}
\label{eq:combined}
\text{Score}_{comb-f}^{(i)} = \left| \alpha \cdot\text{Score}_{ins-f}^{(i)} - (1 - \alpha) \cdot\text{Score}_{del-f}^{(i)} \right|
\end{equation}

This combined score effectively merges the impact of both insertion and deletion methods in determining the most important feature.
By leveraging the $\alpha$ value as a hyperparameter, we are able to determine the optimal balance between the two methods for the best explanation performance.
Insertion and deletion methods are ran in parallel utilizing this final combined score for both methods in the subsequent steps. 

\begin{figure}[ht]
  \centering
  \includegraphics[width=\linewidth]{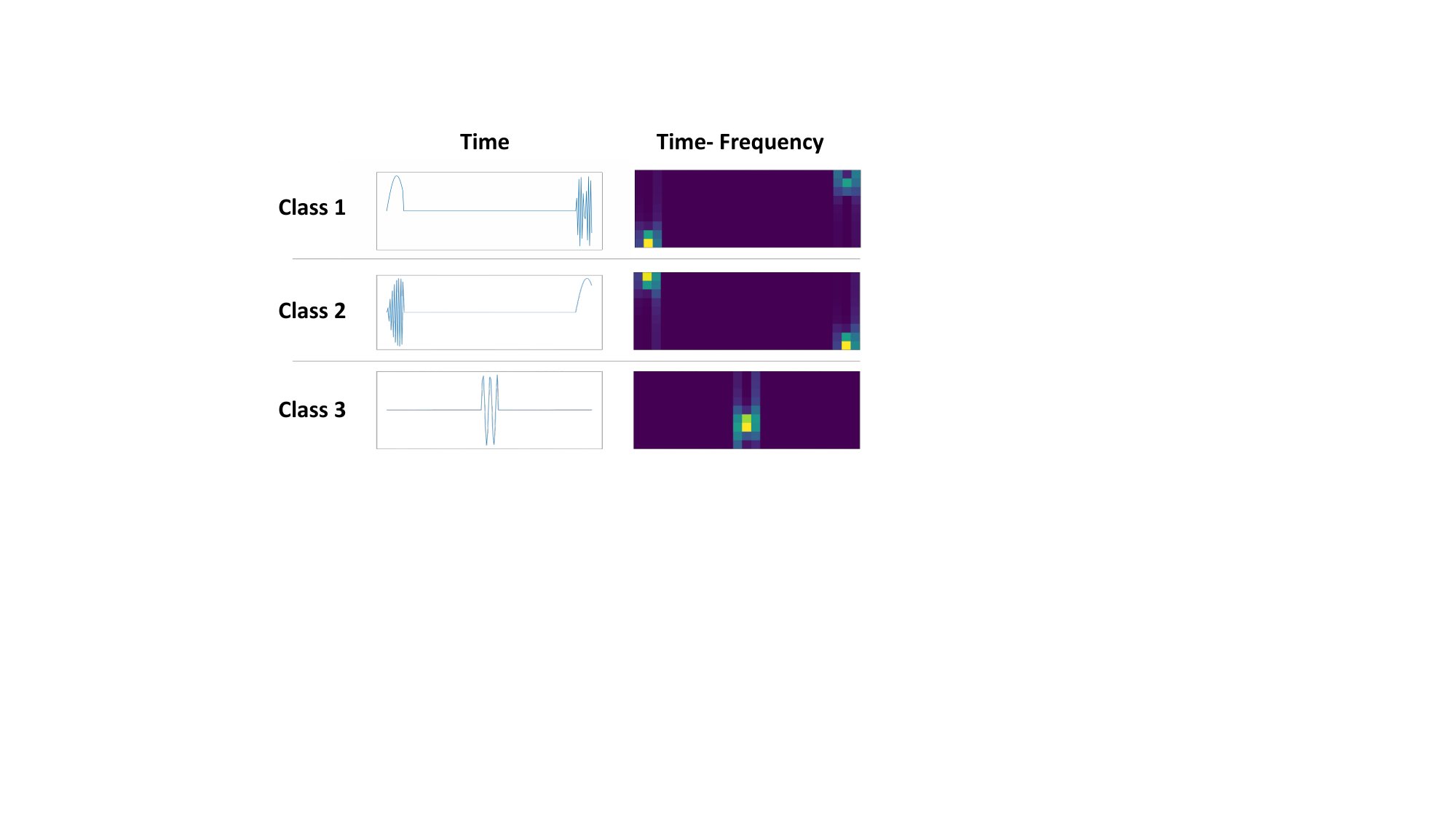}
  \caption{Samples from synthetic dataset for each class. Time and Time-Frequency representation of each sample.}
    \Description{Examples of the synthetic dataset which shows the difference in important regions between classes in the time-domain and the time-frequency domain.}
  \label{samples}
\end{figure}
\setlength{\intextsep}{5pt}

\section{Experiments}
In this section, we describe the experimental settings and datasets of our work. Then, we present and analyze the experimental results. 
We first train the black-box classifiers on the time-series datasets, achieving exceptionally high
accuracy, precision, recall, and F1 scores for most models and datasets
\footnote{We intentionally use time-series datasets that allow near-perfect classification, so we can safely focus on model explanation without having to worry about model training.}.
Then, we use baseline perturbation methods and FIA in the time-domain to collect the top-1 important time-segments.
These segments are collected to compare with important time-frequency features, which are determined with the SpectralX framework.
This comparison experiment is conducted to determine the optimal domain for time-series explanation with average Faithfulness, which measures the average decrease in class probability of the black-box classifier when the most important feature is removed for each class, and average Robustness, which is the difference in explanation before and after perturbation with randomly generated noise. 

Afterwards, given the domain that demonstrated superior performance, we conduct extensive experiments in order to further compare the explanation performance of the baseline perturbation methods and FIA.
In the domain of superior performance, we first generate a synthetic dataset to compare each method's predicted explanations to the ground-truth explanations. 
Then, for real-world time-series datasets, the average Faithfulness@k, which measures the average Faithfulness value for $k$ different top-ranked features, is measured.
This shows the consistency of explanations for the top-$k$ importance rankings from each method. 

Finally, we conduct human evaluation with twenty graduate students studying machine learning to rank the explanation quality between FIA and other perturbation methods for each class of seven different time-series datasets.
We instruct the users to rank the explanation that captures the essence of each target class.  

\subsection{Experimental Settings}
We used three different black-box classifiers: 2-layer bi-directional LSTM, 1D ResNet-34, and 2-layer Transformer model.
For all methods we used batch size of 64 and learning rate of 2e-4.
The Transformer model contained Transformer-encoder layers with model dimension of 64, feedforward dimension 256, and 8 self-attention heads.
The dimension of model was also 64 for LSTM.
For the configuration of baseline and FIA, $P$, which is the number of perturbations, is set to 2000.
For RISE and FIA, $R$, which is the number of masked regions, is set to 10. 
Finally, for the combined method in FIA, the insertion weight was set to 0.2 ($\alpha$) and deletion weight was set to 0.8 ($1 - \alpha$).
For STFT and ISTFT we used window size of 16, hop size of 8, and the Hann window.
All experiments were conducted with NVIDIA GeForce RTX 3090 GPUs.

\subsection{Datasets}
We generate a synthetic dataset in which the discriminative time-frequency features are known and considered as ground-truth features.
As shown in Figure \ref{samples}, the synthetic dataset contains three classes, with each class representing a different frequency range in each time position. Specifically, class 1 samples contain low frequency with less than two oscillations in the first 16 timesteps, and high frequency in the last 16 timesteps with more than 10 oscillations. Class 2 samples have high frequency in the first 16 timesteps, and low frequency in the last 16 timesteps. Class 3 samples contain mid frequency in the mid 16 timesteps with more than four and less than 10 oscillations. Important regions in the time-frequency domain are all different, but the important regions in the time domain overlap between classes 1 and 2. For samples in each class, random Gaussian noise were added to base sinusoidal signal for variability across different samples while maintaining frequency. 1000 samples were generated for each class to create total of 3000 samples.  

We also conduct experiments on nine different datasets of UCR repository \cite{UCRArchive}.
The datasets consist of mixture of multi-class and binary-class datasets, which are: 
CincECGTorso, Twopatterns, Mixedshapes, Arrowhead, Strawbery, Yoga, Ford A, Ford B, and GunpointMaleFemale.
We combine the train and test splits from the synthetic dataset and UCR repository datasets and randomly split the datasets into train (80\%), validation (10\%), and test (10\%) sets. 

\subsection{Baselines}
We consider the following four baseline perturbation methods for integration into SpectralX and comparison to FIA:

First, LIME \cite{ribeiro2016should}, whose flexibility and wide applicability makes it a standard for assessing the local fidelity of explanations.
Second, KernelSHAP, a specific implementation of SHAP that uses a kernel-weighted local regression to estimate Shapley values, which employs efficient optimization and approximation of SHAP values.
Third, RISE \cite{petsiuk2018rise}, whose simplicity and effectiveness in identifying important features in the image-domain, serves as a pertinent baseline to highlight critical features in the time-frequency domain, which contains many similarities to the image-domain.
Finally, we compare LIMESegment \cite{sivill2022limesegment} in the time-domain, which is the current state-of-the-art method for model-agnostic time-series local explanations.

\subsection{Time and Time-Frequency Domain}
In Table \ref{tab:faithfulness_comparison}, we present the average Faithfulness results with standard deviations in both time and time-frequency domains for baseline perturbation methods and FIA. 
When determining the most important explanations, we consider a group of timesteps as a single feature, also known as super-segments.
we fix the number of timesteps to 16 for each super-segment in the time-domain, which is the same as the window size used during STFT for time-frequency domain conversion.

Note that perturbation in the time-domain has an inherent advantage over perturbation in the time-frequency domain in influencing the model output (\textit{i.e.} class probability).
This is because during perturbation, the entire super-segment is masked to zero in the former, but only a specific frequency range is masked in the latter. 
Therefore, the former always makes a larger change in the original signal than the latter.
Despite this disadvantage, all methods show competitive performance in the time-frequency domain compared to the time domain, meaning that we were able to obtain only the essence of each class by minimally manipulating the original signal.
We hypothesize that this is due to important features from the time-frequency domain capturing the class-specific essential regions better than the important features from the time domain.
With the exception of KernelSHAP, the average Faithfulness in time-frequency domain performs better than time-domain for all perturbation methods.

Table \ref{tab:robustness} shows the average Robustness values with standard deviations in both time and time-frequency domain for all methods.
Gaussian noise is added to raw time-series data in the time domain and to the magnitude of STFT output in the time-frequency domain.
The Robustness scores represent the proportion of unchanged super-segment explanations for the time-domain, and proportion of unchanged time-frequency explanations for the time-frequency domain. 
To calculate the Robustness scores, we used the top-8 features from each method for a clearer comparison of performance across the time and time-frequency domains. Considering fewer than 8 features made it challenging to determine the superior domain.
For all methods, average Robustness outperforms in time-frequency domain compared to the time-domain.
From the average Faithfulness and Robustness results, we can conclude that the time-frequency domain is the better performing domain for time-series explanations.

\begin{figure}[t]
  \centering
  \includegraphics[width=\linewidth]{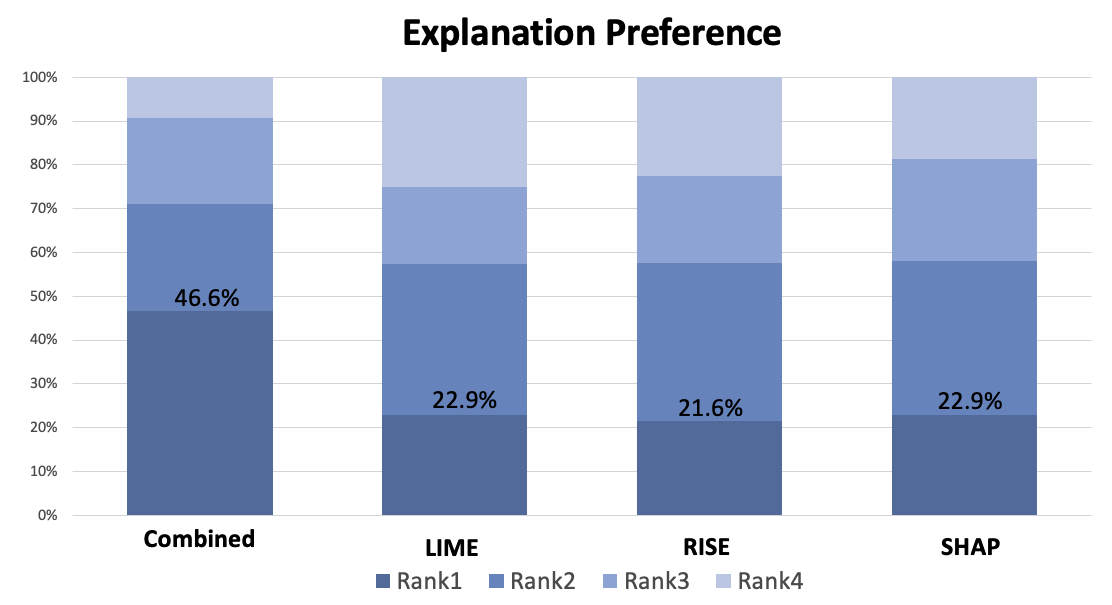}
  \caption{Human evaluation results}
    \Description{Ranks compared between each method from first to fourth place.}
  \label{user_study}
\end{figure}

\subsection{Time-Frequency Evaluations}
As a more direct and objective measure of the accuracy of the predicted explanations, we use the generated synthetic dataset to compare each method's predicted explanations to the ground-truth explanations. The ground-truth explanations are synthesized by the descending magnitude order, from greatest to smallest above zero magnitude.
Figure \ref{samples} shows a sample from each class of the synthetic dataset, where the prominent features in both time and time-frequency domains are displayed. 
The rankings between the ground-truth and predicted explanations are compared with Rank-Biased Overlap (RBO) \cite{webber2010similarity}, which takes into account the rank positions in final score calculations. The formula of RBO is as follows: 
\begin{equation}
RBO(L_1, L_2, d) = (1 - \lambda) \sum_{k=1}^{d} \lambda^{k-1} \frac{|{L_1}_{:k} \cap {L_2}_{:k}|}{k}
\end{equation}
where $L_1$ and $L_2$ are the two ranked feature lists, $d$ represents the depth of the lists to compare, $\lambda$ is the parameter to control the emphasis on the top of the lists, which we set to $0.9$. Although there are many popular ranking methods such as Kendall Tau \cite{kendall1938new} and Spearman's rank correlation coefficient \cite{zar2005spearman}, they do not inherently assign higher weights to top ranks in their calculations. It is crucial in our ground-truth and prediction comparison to take into account the order of the top explanation features, and give more weight to the agreement between ground-truth and predicted at the top of the list, which the RBO is able to do. The RBO calculates a score between 0 and 1, where 0 indicates no overlap and 1 indicates identical rankings. The calculation is based on the cumulative overlap up to each depth $d$ of the explanations, weighted by a factor that decreases the importance of items further down the list. Additionally, we measure the quality of explanations by measuring the proportion of predicted features that are indeed the ground-truth with area under the precision curve (AUP), and measure the portion of ground-truth features identified with area under the recall curve (AUR). For all metrics, higher is better.       
The results are shown in Table \ref{tab:rbo}, where combined method in FIA scores above other methods in most cases. 

We then measure the Faithfulness@k in the nine datasets of the UCR repository. Table \ref{tab:faithfulness@k} presents the average Faithfulness@k that shows the average Faithfulness values for $k$ regions. This metric shows the consistency of explanations not only in the top-1, but also up to the top-8 explanations. The combined method in FIA shows best performance in most $k$ values except for when $k$ is 2. This indicates its superiority in various explanation granularity.      
\renewcommand{\arraystretch}{0.9}
\begin{table}
  \caption{Average Faithfulness comparison for different $\alpha$ values in Time and Time-Frequency Domain for all nine UCR datasets. Mean and standard deviation are shown across three classifiers. The \textbf{boldface} highlights the best performance in each domain.}
  \setlength{\tabcolsep}{10pt}
  \label{tab:alpha_comparison}
  \begin{tabular}{ccc}
    \toprule
    \multicolumn{3}{c}{Average Faithfulness} \\
    \hline\hline 
    $\alpha$ & Time & Time-Frequency \\
    \hline
    $0.8$ & $0.115 \pm 0.06$ & $0.122 \pm 0.06$ \\
    $0.5$ & $0.131 \pm 0.06$ & $0.133 \pm 0.05$ \\
    $0.2$ & $\bf 0.147 \pm 0.04$ & $\bf 0.154 \pm 0.08$ \\
    \hline
    \bottomrule
  \end{tabular}
\end{table}

{\small
\renewcommand{\arraystretch}{0.95}
\begin{table*}[!t]
  \caption{Average Faithfulness@k values for different methods and window/hop sizes in the CinCECGTorso dataset. Mean and standard deviation are across three classifiers. \textbf{Boldface} highlights the best performance for each @k and window/hop size.}\label{tab:faithfulness@k_winhop}
  \setlength{\tabcolsep}{10pt}
  \begin{tabular}{lcccccc}
  \toprule
    & & \multicolumn{5}{c}{Faithfulness@k} \\
    \cmidrule(lr){3-7}
    & Win./Hop & @1 & @2 & @4 & @6 & @8 \\
  \midrule
  \hline
  \textit{Baselines} & & & & & & \\
  \hline
  \multirow{2}{*}{LIME} 
    & 64/32  & \textbf{0.064 $\pm$ 0.04} & 0.098 $\pm$ 0.08 & 0.291 $\pm$ 0.09 & 0.577 $\pm$ 0.03 & 0.688 $\pm$ 0.05 \\
    & 128/64 & 0.155 $\pm$ 0.02 & \textbf{0.340 $\pm$ 0.06} & 0.508 $\pm$ 0.03 & 0.681 $\pm$ 0.02 & 0.810 $\pm$ 0.07 \\
  \hline
  \multirow{2}{*}{KernelSHAP} 
    & 64/32  & 0.059 $\pm$ 0.06 & 0.104 $\pm$ 0.07 & 0.295 $\pm$ 0.08 & 0.541 $\pm$ 0.06 & 0.653 $\pm$ 0.07 \\
    & 128/64 & 0.150 $\pm$ 0.01 & 0.332 $\pm$ 0.04 & 0.502 $\pm$ 0.04 & 0.695 $\pm$ 0.04 & 0.834 $\pm$ 0.05 \\
  \hline
  \multirow{2}{*}{RISE} 
    & 64/32  & 0.062 $\pm$ 0.08 & 0.092 $\pm$ 0.08 & 0.274 $\pm$ 0.09 & 0.562 $\pm$ 0.09 & 0.671 $\pm$ 0.07 \\
    & 128/64 & 0.151 $\pm$ 0.02 & 0.328 $\pm$ 0.06 & 0.496 $\pm$ 0.02 & 0.674 $\pm$ 0.04 & 0.818 $\pm$ 0.06 \\
  \midrule
  \textit{FIA (Ours)} & & & & & & \\
  \hline
  \multirow{2}{*}{Insertion} 
    & 64/32  & 0.055 $\pm$ 0.09 & 0.101 $\pm$ 0.06 & 0.291 $\pm$ 0.05 & 0.560 $\pm$ 0.06 & 0.647 $\pm$ 0.04 \\
    & 128/64 & 0.148 $\pm$ 0.01 & 0.325 $\pm$ 0.05 & 0.485 $\pm$ 0.03 & 0.659 $\pm$ 0.05 & 0.802 $\pm$ 0.04 \\
  \hline
  \multirow{2}{*}{Deletion} 
    & 64/32  & 0.062 $\pm$ 0.06 & 0.105 $\pm$ 0.07 & \textbf{0.313 $\pm$ 0.06} & 0.575 $\pm$ 0.05 & 0.683 $\pm$ 0.08 \\
    & 128/64 & 0.153 $\pm$ 0.02 & 0.336 $\pm$ 0.06 & 0.514 $\pm$ 0.05 & 0.703 $\pm$ 0.07 & 0.842 $\pm$ 0.03 \\
  \hline
  \multirow{2}{*}{Combined} 
    & 64/32  & 0.060 $\pm$ 0.05 & \textbf{0.108 $\pm$ 0.05} & 0.308 $\pm$ 0.08 & \textbf{0.583 $\pm$ 0.04} & \textbf{0.691 $\pm$ 0.06} \\
    & 128/64 & \textbf{0.159 $\pm$ 0.02} & 0.332 $\pm$ 0.03 & \textbf{0.524 $\pm$ 0.04} & \textbf{0.725 $\pm$ 0.04} & \textbf{0.855 $\pm$ 0.04} \\
  \hline
  \bottomrule
  \end{tabular}
\end{table*}
}

\subsection{User Study}
We conducted a user study to assess how well FIA combined method captures class-wise explanations for UCR time-series datasets \cite{UCRArchive}. Each question presented a selection of options featuring plots of class average values of each dataset, and the top-1 important feature produced by each method in the time-frequency domain masked and reverted to the time-domain. Explanation samples are shown in Appendix~\ref{apd:explanation_samples}.
Figure \ref{user_study} presents the ranking results of the explanation preference for each method across a total of seven datasets and 19 classes. We excluded the Mixedshapes and Ford B datasets from the user study because the important features for each class in these datasets were very difficult to distinguish, and including them would have introduced significant noise into the user study. Users ranked the explanation preference of each region highlighted by the four methods from first to fourth place. Ties were allowed for multiple choices. The combined method achieved the highest rank 1 proportion of $46.6\%$, followed by LIME and SHAP which were tied at $22.9\%$ and then RISE with  rank 1 ratio of $21.6\%$.    

\subsection{Ablation Study}
We explore the $\alpha$ values for FIA combined method, and various window and hop sizes for STFT and ISTFT. 
In Table \ref{tab:alpha_comparison}, lower $\alpha$ values for combined method indicate less weight towards the insertion method, and more weight towards the deletion method according to Equation \eqref{eq:combined}. The results show that decrease in the insertion weight and increase in the deletion weight increases the average faithfulness values for both time and time-frequency domains. However, $\alpha$ value of $0.2$ outperforms $\alpha$ value of $0$ (deletion method) as shown in Table \ref{tab:faithfulness@k}. We hypothesize a slight bias towards the deletion method without completely losing the complementary insights from the insertion method produces the optimal synergy that can leverage the strengths of both deletion and insertion methods.   

We observe average Faithfulness@k performance of various methods for different window and hop size values during STFT. We conduct experiments on the CinCECGTorso dataset, which contains the longest time length samples among the nine different datasets. The other datasets are too short to provide meaningful performance evaluations with gradual increase in window and hop size. This variation in window and hop sizes aims to demonstrate that the performance remains relatively consistent with the fixed window and hop size hyperparameters used in the previous experiments, with the combined method showing optimal performance in most @k values. The results are presented in Table \ref{tab:faithfulness@k_winhop}.  

\section{Discussion and Future Work}
There are some limitations in our experiments. First, in SpectralX, we did not explore the optimal hyperparameters in-depth for the STFT window function (we used Hann window in all our experiments), which can possibly alter performance depending on the time-series dataset. Second, FIA provides class-specific explanations for the target class but lacks detailed explanations for individual samples (compared to local models such as LIME).
We also acknowledge that FIA is a generic XAI method capable of being applied across a broader range of domains beyond TSC.
This is a promising avenue for future research, therefore as future work we propose to extend FIA to other domains such as Natural Language Processing or Computer Vision to observe the efficacy in a broader context.        

\section{Conclusion}
In this work, we present a time-frequency based explanation framework, SpectralX, for time-series black-box model explanations. In addition, we propose FIA, which are insertion, deletion, and combined methods. We first compare the performance of various methods in the time and time-frequency domain to show the advantages of time-frequency based explanations, and then compared our proposed combined method in FIA with other baseline methods in the time-frequency domain. By conducting rigorous quantitative evaluations with diverse metrics and a user study to demonstrate the efficacy of our framework, our combined method in FIA showed superior performance compared to other baseline methods.    
\section{Acknowledgements}
This work was supported by the Institute of Information \& Communications Technology Planning \& Evaluation (IITP) grant (No.RS-2019-II190075, No.RS-2022-II220984), and National Research Foundation of Korea (NRF) grant (NRF-2020H1D3A2A03100945), funded by the Korea government (MSIT).

\clearpage
\bibliographystyle{ACM-Reference-Format}

\bibliography{custom}


\begin{thebibliography}{39}


\ifx \showCODEN    \undefined \def \showCODEN     #1{\unskip}     \fi
\ifx \showDOI      \undefined \def \showDOI       #1{#1}\fi
\ifx \showISBNx    \undefined \def \showISBNx     #1{\unskip}     \fi
\ifx \showISBNxiii \undefined \def \showISBNxiii  #1{\unskip}     \fi
\ifx \showISSN     \undefined \def \showISSN      #1{\unskip}     \fi
\ifx \showLCCN     \undefined \def \showLCCN      #1{\unskip}     \fi
\ifx \shownote     \undefined \def \shownote      #1{#1}          \fi
\ifx \showarticletitle \undefined \def \showarticletitle #1{#1}   \fi
\ifx \showURL      \undefined \def \showURL       {\relax}        \fi
\providecommand\bibfield[2]{#2}
\providecommand\bibinfo[2]{#2}
\providecommand\natexlab[1]{#1}
\providecommand\showeprint[2][]{arXiv:#2}

\bibitem[Adadi and Berrada(2018)]%
        {adadi2018peeking}
\bibfield{author}{\bibinfo{person}{Amina Adadi} {and} \bibinfo{person}{Mohammed Berrada}.} \bibinfo{year}{2018}\natexlab{}.
\newblock \showarticletitle{Peeking inside the black-box: a survey on explainable artificial intelligence (XAI)}.
\newblock \bibinfo{journal}{\emph{IEEE access}}  \bibinfo{volume}{6} (\bibinfo{year}{2018}), \bibinfo{pages}{52138--52160}.
\newblock


\bibitem[Bach et~al\mbox{.}(2015)]%
        {bach2015pixel}
\bibfield{author}{\bibinfo{person}{Sebastian Bach}, \bibinfo{person}{Alexander Binder}, \bibinfo{person}{Gr{\'e}goire Montavon}, \bibinfo{person}{Frederick Klauschen}, \bibinfo{person}{Klaus-Robert M{\"u}ller}, {and} \bibinfo{person}{Wojciech Samek}.} \bibinfo{year}{2015}\natexlab{}.
\newblock \showarticletitle{On pixel-wise explanations for non-linear classifier decisions by layer-wise relevance propagation}.
\newblock \bibinfo{journal}{\emph{PloS one}} \bibinfo{volume}{10}, \bibinfo{number}{7} (\bibinfo{year}{2015}), \bibinfo{pages}{e0130140}.
\newblock


\bibitem[Chen et~al\mbox{.}(2015)]%
        {UCRArchive}
\bibfield{author}{\bibinfo{person}{Yanping Chen}, \bibinfo{person}{Eamonn Keogh}, \bibinfo{person}{Bing Hu}, \bibinfo{person}{Nurjahan Begum}, \bibinfo{person}{Anthony Bagnall}, \bibinfo{person}{Abdullah Mueen}, {and} \bibinfo{person}{Gustavo Batista}.} \bibinfo{year}{2015}\natexlab{}.
\newblock \bibinfo{title}{The UCR Time Series Classification Archive}.
\newblock
\newblock
\newblock
\shownote{\url{www.cs.ucr.edu/~eamonn/time_series_data/}}.


\bibitem[Crabb{\'e} and Van Der~Schaar(2021)]%
        {crabbe2021explaining}
\bibfield{author}{\bibinfo{person}{Jonathan Crabb{\'e}} {and} \bibinfo{person}{Mihaela Van Der~Schaar}.} \bibinfo{year}{2021}\natexlab{}.
\newblock \showarticletitle{Explaining time series predictions with dynamic masks}. In \bibinfo{booktitle}{\emph{International Conference on Machine Learning}}. PMLR, \bibinfo{pages}{2166--2177}.
\newblock


\bibitem[Duell et~al\mbox{.}(2021)]%
        {duell2021comparison}
\bibfield{author}{\bibinfo{person}{Jamie Duell}, \bibinfo{person}{Xiuyi Fan}, \bibinfo{person}{Bruce Burnett}, \bibinfo{person}{Gert Aarts}, {and} \bibinfo{person}{Shang-Ming Zhou}.} \bibinfo{year}{2021}\natexlab{}.
\newblock \showarticletitle{A comparison of explanations given by explainable artificial intelligence methods on analysing electronic health records}. In \bibinfo{booktitle}{\emph{2021 IEEE EMBS International Conference on Biomedical and Health Informatics (BHI)}}. IEEE, \bibinfo{pages}{1--4}.
\newblock


\bibitem[Fong and Vedaldi(2017)]%
        {fong2017interpretable}
\bibfield{author}{\bibinfo{person}{Ruth~C Fong} {and} \bibinfo{person}{Andrea Vedaldi}.} \bibinfo{year}{2017}\natexlab{}.
\newblock \showarticletitle{Interpretable explanations of black boxes by meaningful perturbation}. In \bibinfo{booktitle}{\emph{Proceedings of the IEEE international conference on computer vision}}. \bibinfo{pages}{3429--3437}.
\newblock


\bibitem[Gramegna and Giudici(2021)]%
        {gramegna2021shap}
\bibfield{author}{\bibinfo{person}{Alex Gramegna} {and} \bibinfo{person}{Paolo Giudici}.} \bibinfo{year}{2021}\natexlab{}.
\newblock \showarticletitle{SHAP and LIME: an evaluation of discriminative power in credit risk}.
\newblock \bibinfo{journal}{\emph{Frontiers in Artificial Intelligence}}  \bibinfo{volume}{4} (\bibinfo{year}{2021}), \bibinfo{pages}{752558}.
\newblock


\bibitem[Guillemé et~al\mbox{.}(2019)]%
        {8995349}
\bibfield{author}{\bibinfo{person}{Maël Guillemé}, \bibinfo{person}{Véronique Masson}, \bibinfo{person}{Laurence Rozé}, {and} \bibinfo{person}{Alexandre Termier}.} \bibinfo{year}{2019}\natexlab{}.
\newblock \showarticletitle{Agnostic Local Explanation for Time Series Classification}. In \bibinfo{booktitle}{\emph{2019 IEEE 31st International Conference on Tools with Artificial Intelligence (ICTAI)}}. \bibinfo{pages}{432--439}.
\newblock
\urldef\tempurl%
\url{https://doi.org/10.1109/ICTAI.2019.00067}
\showDOI{\tempurl}


\bibitem[Harris(1978)]%
        {harris1978use}
\bibfield{author}{\bibinfo{person}{Fredric~J Harris}.} \bibinfo{year}{1978}\natexlab{}.
\newblock \showarticletitle{On the use of windows for harmonic analysis with the discrete Fourier transform}.
\newblock \bibinfo{journal}{\emph{Proc. IEEE}} \bibinfo{volume}{66}, \bibinfo{number}{1} (\bibinfo{year}{1978}), \bibinfo{pages}{51--83}.
\newblock


\bibitem[Hochreiter and Schmidhuber(1997)]%
        {hochreiter1997long}
\bibfield{author}{\bibinfo{person}{Sepp Hochreiter} {and} \bibinfo{person}{J{\"u}rgen Schmidhuber}.} \bibinfo{year}{1997}\natexlab{}.
\newblock \showarticletitle{Long short-term memory}.
\newblock \bibinfo{journal}{\emph{Neural computation}} \bibinfo{volume}{9}, \bibinfo{number}{8} (\bibinfo{year}{1997}), \bibinfo{pages}{1735--1780}.
\newblock


\bibitem[Hsieh et~al\mbox{.}(2021)]%
        {hsieh2021explainable}
\bibfield{author}{\bibinfo{person}{Tsung-Yu Hsieh}, \bibinfo{person}{Suhang Wang}, \bibinfo{person}{Yiwei Sun}, {and} \bibinfo{person}{Vasant Honavar}.} \bibinfo{year}{2021}\natexlab{}.
\newblock \showarticletitle{Explainable multivariate time series classification: a deep neural network which learns to attend to important variables as well as time intervals}. In \bibinfo{booktitle}{\emph{Proceedings of the 14th ACM international conference on web search and data mining}}. \bibinfo{pages}{607--615}.
\newblock


\bibitem[Ivanovs et~al\mbox{.}(2021)]%
        {ivanovs2021perturbation}
\bibfield{author}{\bibinfo{person}{Maksims Ivanovs}, \bibinfo{person}{Roberts Kadikis}, {and} \bibinfo{person}{Kaspars Ozols}.} \bibinfo{year}{2021}\natexlab{}.
\newblock \showarticletitle{Perturbation-based methods for explaining deep neural networks: A survey}.
\newblock \bibinfo{journal}{\emph{Pattern Recognition Letters}}  \bibinfo{volume}{150} (\bibinfo{year}{2021}), \bibinfo{pages}{228--234}.
\newblock


\bibitem[Jalali et~al\mbox{.}(2020)]%
        {jalali2020machine}
\bibfield{author}{\bibinfo{person}{Anahid Jalali}, \bibinfo{person}{Alexander Schindler}, \bibinfo{person}{Bernhard Haslhofer}, {and} \bibinfo{person}{Andreas Rauber}.} \bibinfo{year}{2020}\natexlab{}.
\newblock \showarticletitle{Machine learning interpretability techniques for outage prediction: a comparative study}. In \bibinfo{booktitle}{\emph{PHM Society European Conference}}, Vol.~\bibinfo{volume}{5}. \bibinfo{pages}{10--10}.
\newblock


\bibitem[Karim et~al\mbox{.}(2017)]%
        {karim2017lstm}
\bibfield{author}{\bibinfo{person}{Fazle Karim}, \bibinfo{person}{Somshubra Majumdar}, \bibinfo{person}{Houshang Darabi}, {and} \bibinfo{person}{Shun Chen}.} \bibinfo{year}{2017}\natexlab{}.
\newblock \showarticletitle{LSTM fully convolutional networks for time series classification}.
\newblock \bibinfo{journal}{\emph{IEEE access}}  \bibinfo{volume}{6} (\bibinfo{year}{2017}), \bibinfo{pages}{1662--1669}.
\newblock


\bibitem[Kemp et~al\mbox{.}(2000)]%
        {kemp2000analysis}
\bibfield{author}{\bibinfo{person}{Bob Kemp}, \bibinfo{person}{Aeilko~H Zwinderman}, \bibinfo{person}{Bert Tuk}, \bibinfo{person}{Hilbert~AC Kamphuisen}, {and} \bibinfo{person}{Josefien~JL Oberye}.} \bibinfo{year}{2000}\natexlab{}.
\newblock \showarticletitle{Analysis of a sleep-dependent neuronal feedback loop: the slow-wave microcontinuity of the EEG}.
\newblock \bibinfo{journal}{\emph{IEEE Transactions on Biomedical Engineering}} \bibinfo{volume}{47}, \bibinfo{number}{9} (\bibinfo{year}{2000}), \bibinfo{pages}{1185--1194}.
\newblock


\bibitem[Kendall(1938)]%
        {kendall1938new}
\bibfield{author}{\bibinfo{person}{Maurice~G Kendall}.} \bibinfo{year}{1938}\natexlab{}.
\newblock \showarticletitle{A new measure of rank correlation}.
\newblock \bibinfo{journal}{\emph{Biometrika}} \bibinfo{volume}{30}, \bibinfo{number}{1/2} (\bibinfo{year}{1938}), \bibinfo{pages}{81--93}.
\newblock


\bibitem[LeCun et~al\mbox{.}(2015)]%
        {lecun2015deep}
\bibfield{author}{\bibinfo{person}{Yann LeCun}, \bibinfo{person}{Yoshua Bengio}, {and} \bibinfo{person}{Geoffrey Hinton}.} \bibinfo{year}{2015}\natexlab{}.
\newblock \showarticletitle{Deep learning}.
\newblock \bibinfo{journal}{\emph{nature}} \bibinfo{volume}{521}, \bibinfo{number}{7553} (\bibinfo{year}{2015}), \bibinfo{pages}{436--444}.
\newblock


\bibitem[Lundberg and Lee(2017)]%
        {lundberg2017unified}
\bibfield{author}{\bibinfo{person}{Scott~M Lundberg} {and} \bibinfo{person}{Su-In Lee}.} \bibinfo{year}{2017}\natexlab{}.
\newblock \showarticletitle{A unified approach to interpreting model predictions}.
\newblock \bibinfo{journal}{\emph{Advances in neural information processing systems}}  \bibinfo{volume}{30} (\bibinfo{year}{2017}).
\newblock


\bibitem[Makridis et~al\mbox{.}(2023)]%
        {makridis2023xai}
\bibfield{author}{\bibinfo{person}{Georgios Makridis}, \bibinfo{person}{Georgios Fatouros}, \bibinfo{person}{Vasileios Koukos}, \bibinfo{person}{Dimitrios Kotios}, \bibinfo{person}{Dimosthenis Kyriazis}, {and} \bibinfo{person}{John Soldatos}.} \bibinfo{year}{2023}\natexlab{}.
\newblock \showarticletitle{XAI for time-series classification leveraging image highlight methods}. In \bibinfo{booktitle}{\emph{International Conference on Management of Digital}}. Springer, \bibinfo{pages}{382--396}.
\newblock


\bibitem[Mishra et~al\mbox{.}(2020)]%
        {mishra2020reliable}
\bibfield{author}{\bibinfo{person}{Saumitra Mishra}, \bibinfo{person}{Emmanouil Benetos}, \bibinfo{person}{Bob~LT Sturm}, {and} \bibinfo{person}{Simon Dixon}.} \bibinfo{year}{2020}\natexlab{}.
\newblock \showarticletitle{Reliable local explanations for machine listening}. In \bibinfo{booktitle}{\emph{2020 International Joint Conference on Neural Networks (IJCNN)}}. IEEE, \bibinfo{pages}{1--8}.
\newblock


\bibitem[Moody(1983)]%
        {moody1983new}
\bibfield{author}{\bibinfo{person}{George Moody}.} \bibinfo{year}{1983}\natexlab{}.
\newblock \showarticletitle{A new method for detecting atrial fibrillation using RR intervals}.
\newblock \bibinfo{journal}{\emph{Proc. Comput. Cardiol.}}  \bibinfo{volume}{10} (\bibinfo{year}{1983}), \bibinfo{pages}{227--230}.
\newblock


\bibitem[Neves et~al\mbox{.}(2021)]%
        {neves2021interpretable}
\bibfield{author}{\bibinfo{person}{Ines Neves}, \bibinfo{person}{Duarte Folgado}, \bibinfo{person}{Sara Santos}, \bibinfo{person}{Marilia Barandas}, \bibinfo{person}{Andrea Campagner}, \bibinfo{person}{Luca Ronzio}, \bibinfo{person}{Federico Cabitza}, {and} \bibinfo{person}{Hugo Gamboa}.} \bibinfo{year}{2021}\natexlab{}.
\newblock \showarticletitle{Interpretable heartbeat classification using local model-agnostic explanations on ECGs}.
\newblock \bibinfo{journal}{\emph{Computers in Biology and Medicine}}  \bibinfo{volume}{133} (\bibinfo{year}{2021}), \bibinfo{pages}{104393}.
\newblock


\bibitem[Nguyen et~al\mbox{.}(2018)]%
        {nguyen2018interpretable}
\bibfield{author}{\bibinfo{person}{Thach~Le Nguyen}, \bibinfo{person}{Severin Gsponer}, \bibinfo{person}{Iulia Ilie}, {and} \bibinfo{person}{Georgiana Ifrim}.} \bibinfo{year}{2018}\natexlab{}.
\newblock \showarticletitle{Interpretable time series classification using all-subsequence learning and symbolic representations in time and frequency domains}.
\newblock \bibinfo{journal}{\emph{arXiv preprint arXiv:1808.04022}} (\bibinfo{year}{2018}).
\newblock


\bibitem[Petsiuk et~al\mbox{.}(2018)]%
        {petsiuk2018rise}
\bibfield{author}{\bibinfo{person}{Vitali Petsiuk}, \bibinfo{person}{Abir Das}, {and} \bibinfo{person}{Kate Saenko}.} \bibinfo{year}{2018}\natexlab{}.
\newblock \showarticletitle{Rise: Randomized input sampling for explanation of black-box models}.
\newblock \bibinfo{journal}{\emph{arXiv preprint arXiv:1806.07421}} (\bibinfo{year}{2018}).
\newblock


\bibitem[Raab et~al\mbox{.}(2023)]%
        {raab2023xai4eeg}
\bibfield{author}{\bibinfo{person}{Dominik Raab}, \bibinfo{person}{Andreas Theissler}, {and} \bibinfo{person}{Myra Spiliopoulou}.} \bibinfo{year}{2023}\natexlab{}.
\newblock \showarticletitle{XAI4EEG: spectral and spatio-temporal explanation of deep learning-based seizure detection in EEG time series}.
\newblock \bibinfo{journal}{\emph{Neural Computing and Applications}} \bibinfo{volume}{35}, \bibinfo{number}{14} (\bibinfo{year}{2023}), \bibinfo{pages}{10051--10068}.
\newblock


\bibitem[Ribeiro et~al\mbox{.}(2016)]%
        {ribeiro2016should}
\bibfield{author}{\bibinfo{person}{Marco~Tulio Ribeiro}, \bibinfo{person}{Sameer Singh}, {and} \bibinfo{person}{Carlos Guestrin}.} \bibinfo{year}{2016}\natexlab{}.
\newblock \showarticletitle{" Why should i trust you?" Explaining the predictions of any classifier}. In \bibinfo{booktitle}{\emph{Proceedings of the 22nd ACM SIGKDD international conference on knowledge discovery and data mining}}. \bibinfo{pages}{1135--1144}.
\newblock


\bibitem[Saluja et~al\mbox{.}(2021)]%
        {saluja2021towards}
\bibfield{author}{\bibinfo{person}{Rohit Saluja}, \bibinfo{person}{Avleen Malhi}, \bibinfo{person}{Samanta Knapi{\v{c}}}, \bibinfo{person}{Kary Fr{\"a}mling}, {and} \bibinfo{person}{Cicek Cavdar}.} \bibinfo{year}{2021}\natexlab{}.
\newblock \showarticletitle{Towards a rigorous evaluation of explainability for multivariate time series}.
\newblock \bibinfo{journal}{\emph{arXiv preprint arXiv:2104.04075}} (\bibinfo{year}{2021}).
\newblock


\bibitem[Schlegel et~al\mbox{.}(2019)]%
        {schlegel2019towards}
\bibfield{author}{\bibinfo{person}{Udo Schlegel}, \bibinfo{person}{Hiba Arnout}, \bibinfo{person}{Mennatallah El-Assady}, \bibinfo{person}{Daniela Oelke}, {and} \bibinfo{person}{Daniel~A Keim}.} \bibinfo{year}{2019}\natexlab{}.
\newblock \showarticletitle{Towards a rigorous evaluation of XAI methods on time series}. In \bibinfo{booktitle}{\emph{2019 IEEE/CVF International Conference on Computer Vision Workshop (ICCVW)}}. IEEE, \bibinfo{pages}{4197--4201}.
\newblock


\bibitem[Schlegel and Keim(2023)]%
        {schlegel2023deep}
\bibfield{author}{\bibinfo{person}{Udo Schlegel} {and} \bibinfo{person}{Daniel~A Keim}.} \bibinfo{year}{2023}\natexlab{}.
\newblock \showarticletitle{A Deep Dive into Perturbations as Evaluation Technique for Time Series XAI}. In \bibinfo{booktitle}{\emph{World Conference on Explainable Artificial Intelligence}}. Springer, \bibinfo{pages}{165--180}.
\newblock


\bibitem[Selvaraju et~al\mbox{.}(2017)]%
        {selvaraju2017grad}
\bibfield{author}{\bibinfo{person}{Ramprasaath~R Selvaraju}, \bibinfo{person}{Michael Cogswell}, \bibinfo{person}{Abhishek Das}, \bibinfo{person}{Ramakrishna Vedantam}, \bibinfo{person}{Devi Parikh}, {and} \bibinfo{person}{Dhruv Batra}.} \bibinfo{year}{2017}\natexlab{}.
\newblock \showarticletitle{Grad-cam: Visual explanations from deep networks via gradient-based localization}. In \bibinfo{booktitle}{\emph{Proceedings of the IEEE international conference on computer vision}}. \bibinfo{pages}{618--626}.
\newblock


\bibitem[Sivill and Flach(2022)]%
        {sivill2022limesegment}
\bibfield{author}{\bibinfo{person}{Torty Sivill} {and} \bibinfo{person}{Peter Flach}.} \bibinfo{year}{2022}\natexlab{}.
\newblock \showarticletitle{Limesegment: Meaningful, realistic time series explanations}. In \bibinfo{booktitle}{\emph{International Conference on Artificial Intelligence and Statistics}}. PMLR, \bibinfo{pages}{3418--3433}.
\newblock


\bibitem[Sundararajan et~al\mbox{.}(2017)]%
        {sundararajan2017axiomatic}
\bibfield{author}{\bibinfo{person}{Mukund Sundararajan}, \bibinfo{person}{Ankur Taly}, {and} \bibinfo{person}{Qiqi Yan}.} \bibinfo{year}{2017}\natexlab{}.
\newblock \showarticletitle{Axiomatic attribution for deep networks}. In \bibinfo{booktitle}{\emph{International conference on machine learning}}. PMLR, \bibinfo{pages}{3319--3328}.
\newblock


\bibitem[Theissler et~al\mbox{.}(2022)]%
        {theissler2022explainable}
\bibfield{author}{\bibinfo{person}{Andreas Theissler}, \bibinfo{person}{Francesco Spinnato}, \bibinfo{person}{Udo Schlegel}, {and} \bibinfo{person}{Riccardo Guidotti}.} \bibinfo{year}{2022}\natexlab{}.
\newblock \showarticletitle{Explainable AI for time series classification: a review, taxonomy and research directions}.
\newblock \bibinfo{journal}{\emph{IEEE Access}} (\bibinfo{year}{2022}).
\newblock


\bibitem[Vaswani et~al\mbox{.}(2017)]%
        {vaswani2017attention}
\bibfield{author}{\bibinfo{person}{Ashish Vaswani}, \bibinfo{person}{Noam Shazeer}, \bibinfo{person}{Niki Parmar}, \bibinfo{person}{Jakob Uszkoreit}, \bibinfo{person}{Llion Jones}, \bibinfo{person}{Aidan~N Gomez}, \bibinfo{person}{{\L}ukasz Kaiser}, {and} \bibinfo{person}{Illia Polosukhin}.} \bibinfo{year}{2017}\natexlab{}.
\newblock \showarticletitle{Attention is all you need}.
\newblock \bibinfo{journal}{\emph{Advances in neural information processing systems}}  \bibinfo{volume}{30} (\bibinfo{year}{2017}).
\newblock


\bibitem[Vinayavekhin et~al\mbox{.}(2018)]%
        {vinayavekhin2018focusing}
\bibfield{author}{\bibinfo{person}{Phongtharin Vinayavekhin}, \bibinfo{person}{Subhajit Chaudhury}, \bibinfo{person}{Asim Munawar}, \bibinfo{person}{Don~Joven Agravante}, \bibinfo{person}{Giovanni De~Magistris}, \bibinfo{person}{Daiki Kimura}, {and} \bibinfo{person}{Ryuki Tachibana}.} \bibinfo{year}{2018}\natexlab{}.
\newblock \showarticletitle{Focusing on what is relevant: Time-series learning and understanding using attention}. In \bibinfo{booktitle}{\emph{2018 24th International Conference on Pattern Recognition (ICPR)}}. IEEE, \bibinfo{pages}{2624--2629}.
\newblock


\bibitem[Wang et~al\mbox{.}(2023)]%
        {wang2023wavelet}
\bibfield{author}{\bibinfo{person}{Jingyuan Wang}, \bibinfo{person}{Chen Yang}, \bibinfo{person}{Xiaohan Jiang}, {and} \bibinfo{person}{Junjie Wu}.} \bibinfo{year}{2023}\natexlab{}.
\newblock \showarticletitle{WHEN: A Wavelet-DTW Hybrid Attention Network for Heterogeneous Time Series Analysis}. In \bibinfo{booktitle}{\emph{Proceedings of the 29th ACM SIGKDD Conference on Knowledge Discovery and Data Mining}}. \bibinfo{pages}{2361--2373}.
\newblock


\bibitem[Webber et~al\mbox{.}(2010)]%
        {webber2010similarity}
\bibfield{author}{\bibinfo{person}{William Webber}, \bibinfo{person}{Alistair Moffat}, {and} \bibinfo{person}{Justin Zobel}.} \bibinfo{year}{2010}\natexlab{}.
\newblock \showarticletitle{A similarity measure for indefinite rankings}.
\newblock \bibinfo{journal}{\emph{ACM Transactions on Information Systems (TOIS)}} \bibinfo{volume}{28}, \bibinfo{number}{4} (\bibinfo{year}{2010}), \bibinfo{pages}{1--38}.
\newblock


\bibitem[Wu and Jiang(2023)]%
        {wu2023time}
\bibfield{author}{\bibinfo{person}{Zhou Wu} {and} \bibinfo{person}{Ruiqi Jiang}.} \bibinfo{year}{2023}\natexlab{}.
\newblock \showarticletitle{Time-series benchmarks based on frequency features for fair comparative evaluation}.
\newblock \bibinfo{journal}{\emph{Neural Computing and Applications}} (\bibinfo{year}{2023}), \bibinfo{pages}{1--13}.
\newblock


\bibitem[Zar(2005)]%
        {zar2005spearman}
\bibfield{author}{\bibinfo{person}{Jerrold~H Zar}.} \bibinfo{year}{2005}\natexlab{}.
\newblock \showarticletitle{Spearman rank correlation}.
\newblock \bibinfo{journal}{\emph{Encyclopedia of Biostatistics}}  \bibinfo{volume}{7} (\bibinfo{year}{2005}).
\newblock


\end{thebibliography}

\appendix

\onecolumn 

\section{Explanation Samples}
\label{apd:explanation_samples}

This section presents explanation samples from UCR repository datasets.
As denoted in the legend of each figure, \textbf{sample} represents the explanation sample and \textbf{masking} represents the inverse-STFT reconstructed signal after masking the most important feature from the time-frequency domain spectrogram. The most important feature is determined by each perturbation-based model stated in the title of each figure. The yellow-region represents the difference between the explanation sample and the masked reconstructed signal, which visualizes the effect of the important time-frequency feature.

\subsection{Arrowhead Dataset}
\begin{figure*}[!ht]
    \centering
    \begin{tabular}{@{}ccc@{}}
        \multicolumn{3}{c}{\includegraphics[width=0.48\textwidth]{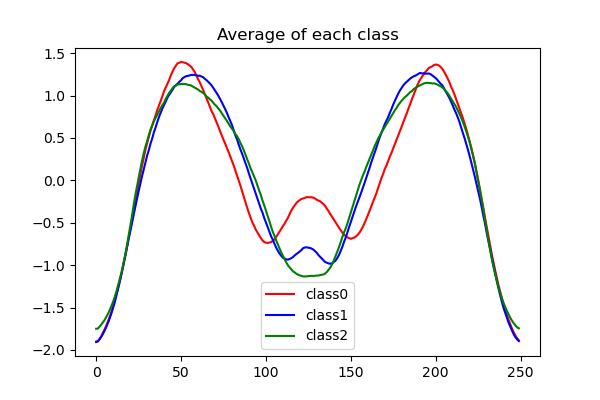}} \\
        \multicolumn{3}{c}{(a) Average of test samples in each class} \\[12pt]
        \includegraphics[width=0.32\textwidth]{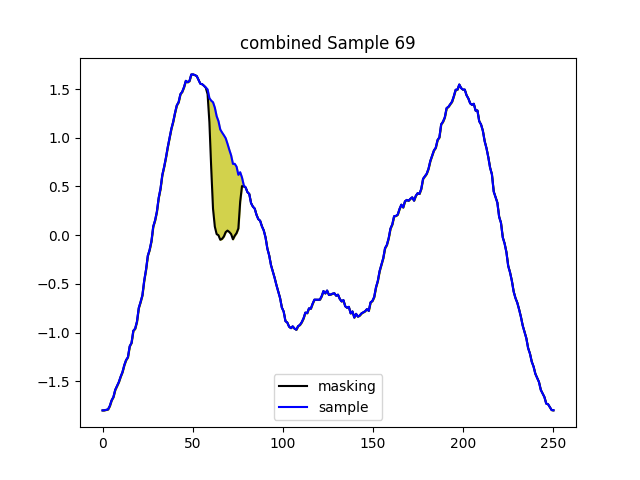} &
        \includegraphics[width=0.32\textwidth]{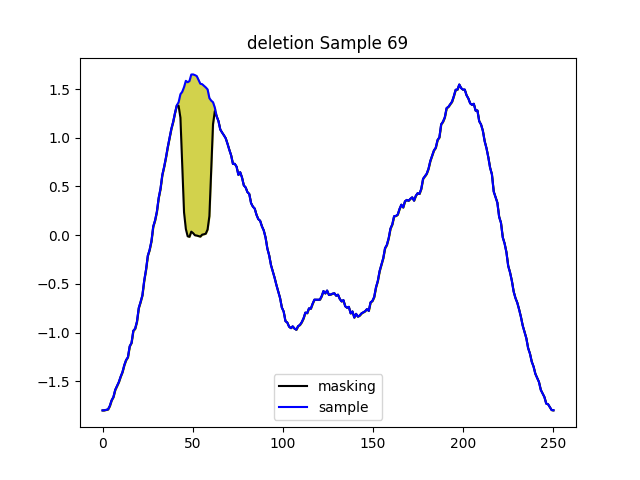} &
        \includegraphics[width=0.32\textwidth]{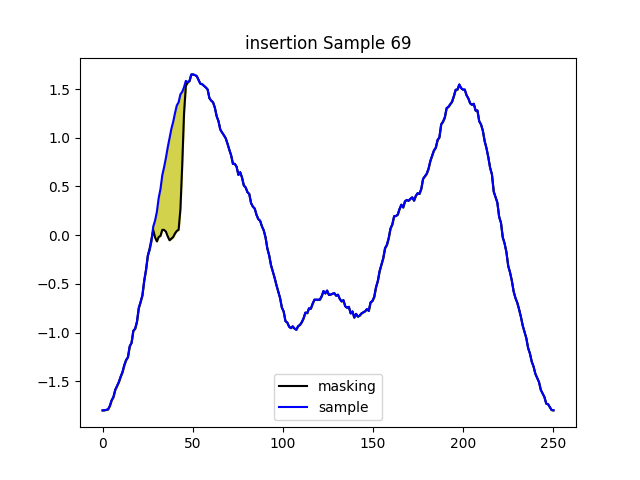} \\
        (b) Combined & (c) Deletion & (d) Insertion \\[12pt]
        \includegraphics[width=0.32\textwidth]{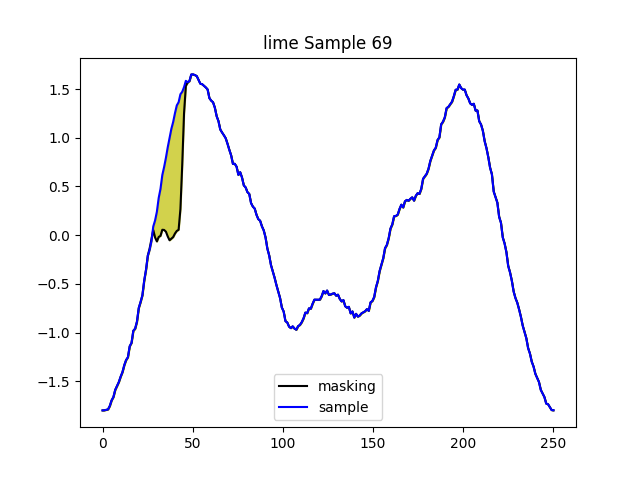} &
        \includegraphics[width=0.32\textwidth]{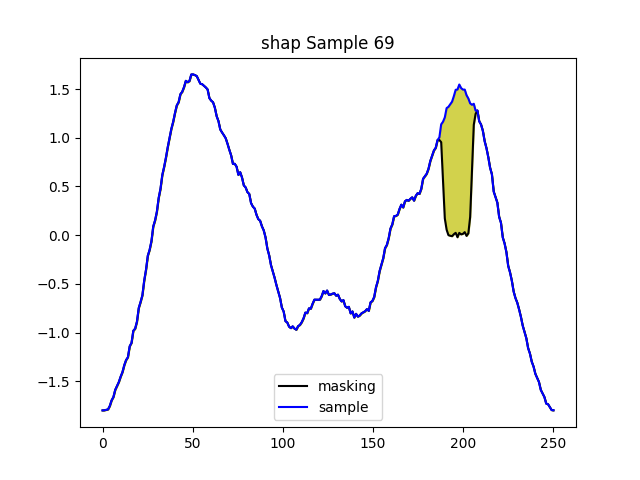} &
        \includegraphics[width=0.32\textwidth]{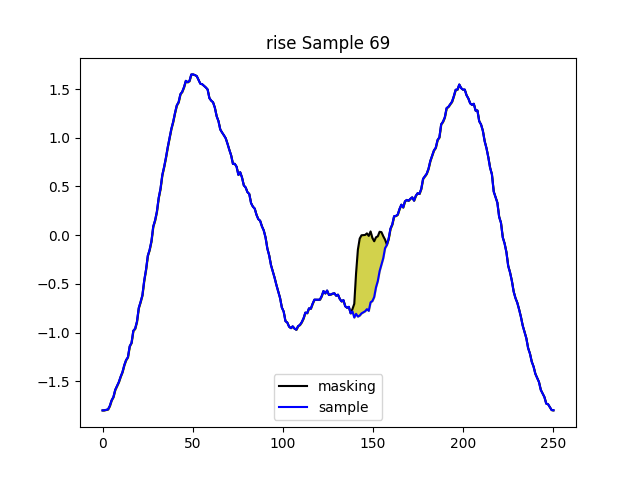} \\
        (e) LIME & (f) SHAP & (g) RISE
    \end{tabular}
    \caption{Explanation samples for Arrowhead dataset, Class 1, Sample 69}
    \label{fig:arrowhead_samples}
\end{figure*}

\clearpage
\subsection{Yoga Dataset}
\begin{figure*}[!ht]
    \centering
    \begin{tabular}{@{}ccc@{}}
        \multicolumn{3}{c}{\includegraphics[width=0.48\textwidth]{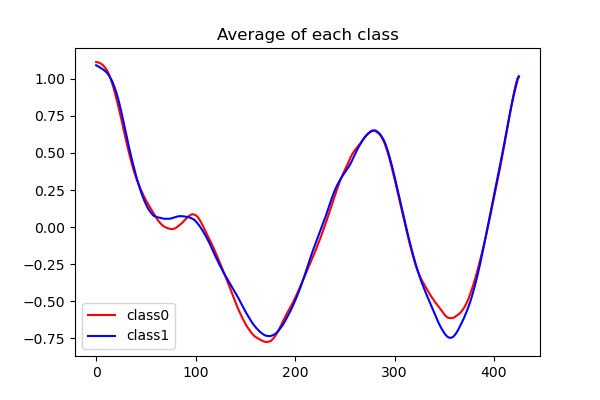}} \\
        \multicolumn{3}{c}{(a) Average of test samples in each class} \\[12pt]
        \includegraphics[width=0.32\textwidth]{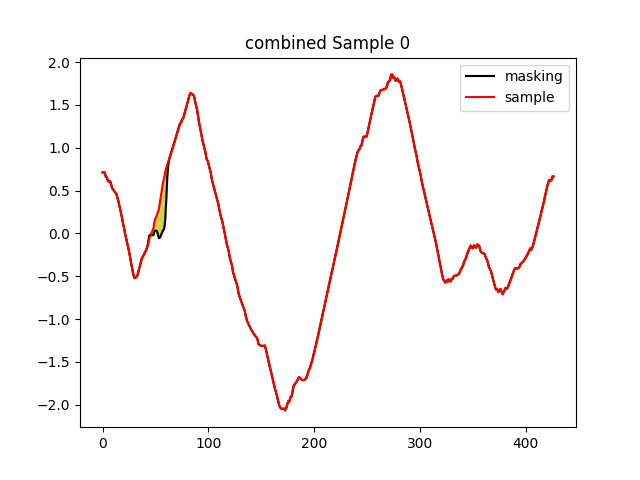} &
        \includegraphics[width=0.32\textwidth]{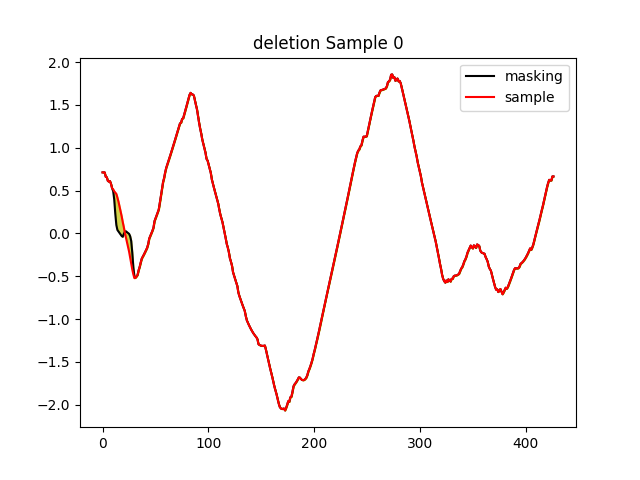} &
        \includegraphics[width=0.32\textwidth]{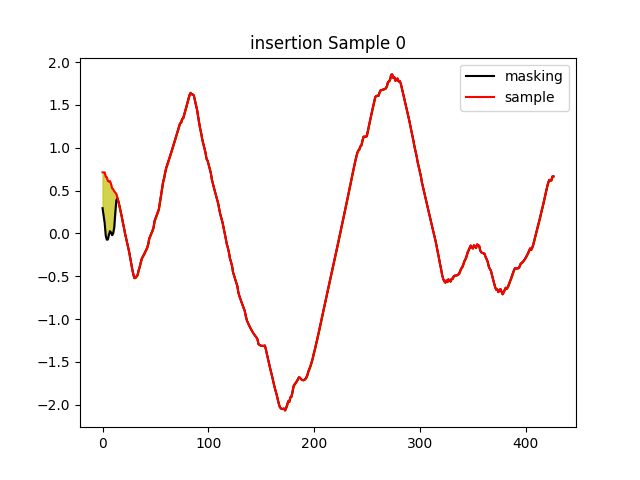} \\
        (b) Combined & (c) Deletion & (d) Insertion \\[12pt]
        \includegraphics[width=0.32\textwidth]{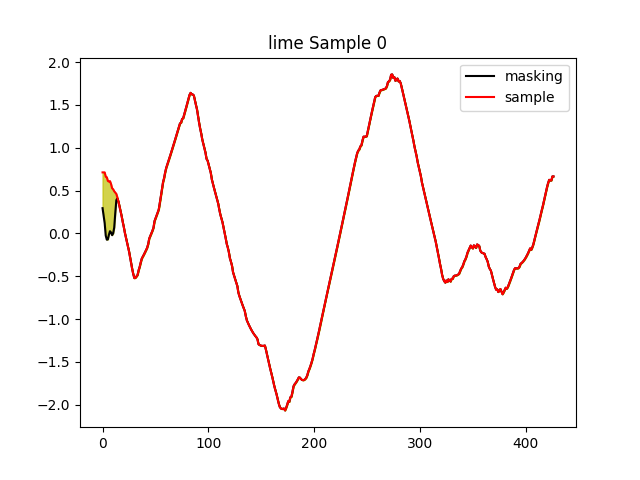} &
        \includegraphics[width=0.32\textwidth]{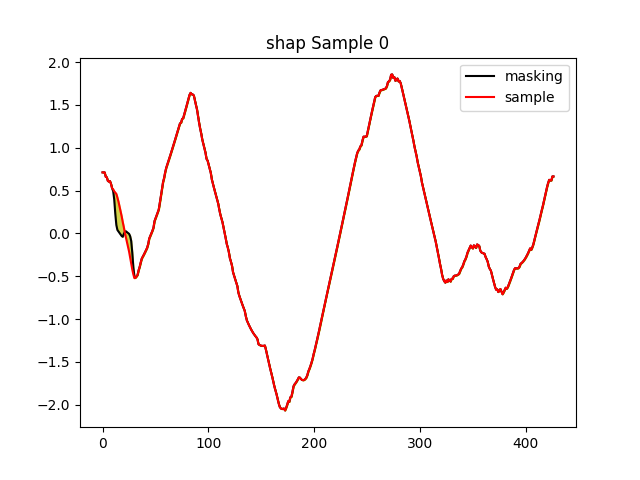} &
        \includegraphics[width=0.32\textwidth]{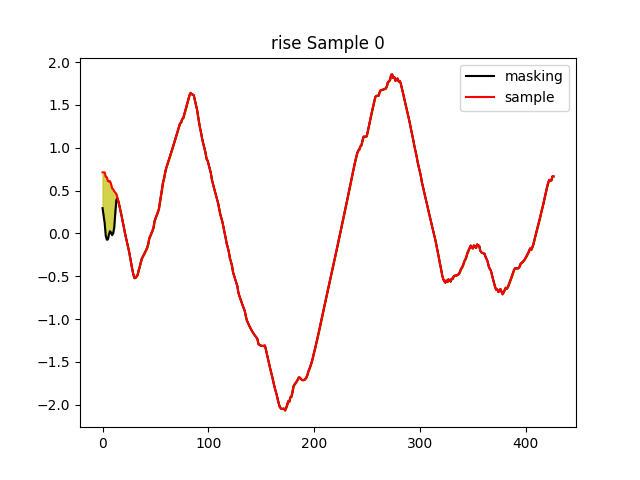} \\
        (e) LIME & (f) SHAP & (g) RISE
    \end{tabular}
    \caption{Explanation samples for Yoga dataset, Class 0, Sample 0}
    \label{fig:yoga_samples_0}
\end{figure*}

\begin{figure*}[t]
    \centering
    \begin{tabular}{@{}ccc@{}}
        \multicolumn{3}{c}{\includegraphics[width=0.48\textwidth]{Explanation_Samples/yoga/Avg.png}} \\
        \multicolumn{3}{c}{(a) Average of test samples in each class} \\[12pt]
        \includegraphics[width=0.32\textwidth]{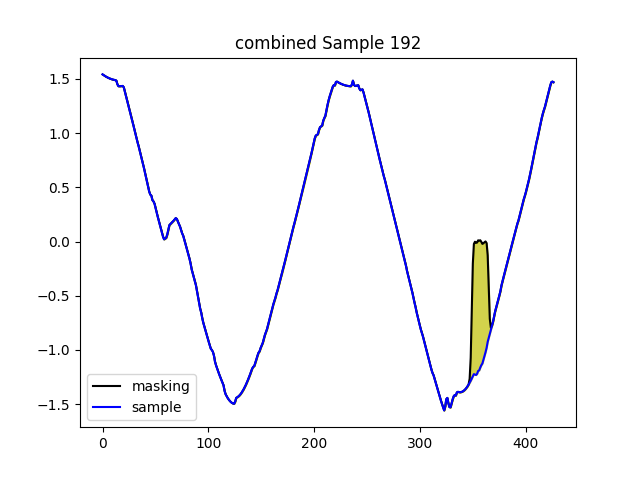} &
        \includegraphics[width=0.32\textwidth]{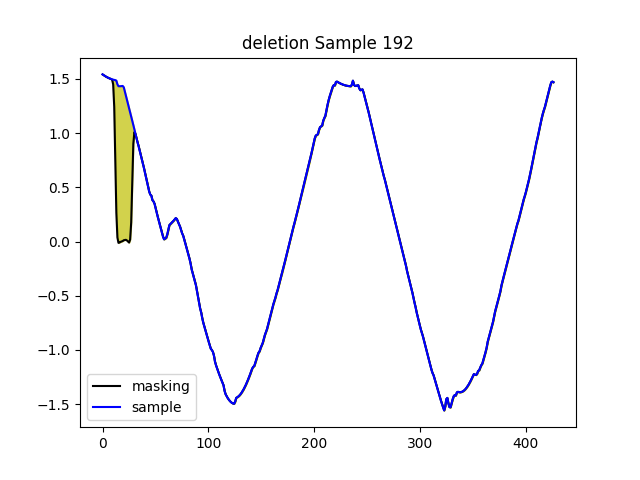} &
        \includegraphics[width=0.32\textwidth]{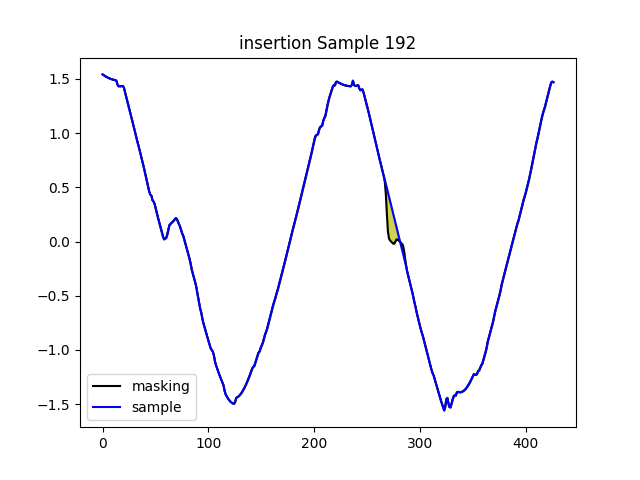} \\
        (b) Combined & (c) Deletion & (d) Insertion \\[12pt]
        \includegraphics[width=0.32\textwidth]{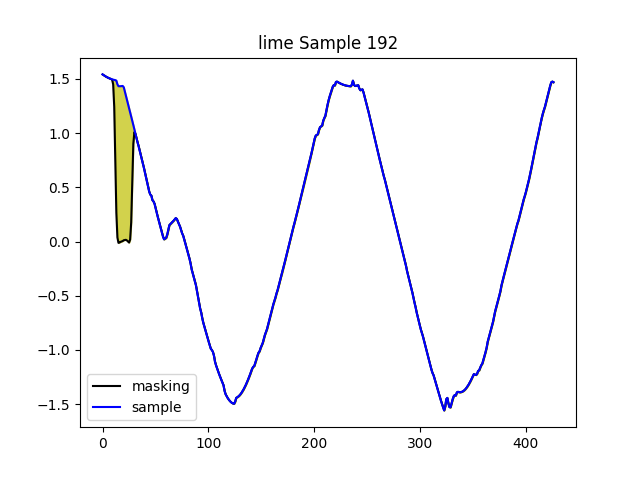} &
        \includegraphics[width=0.32\textwidth]{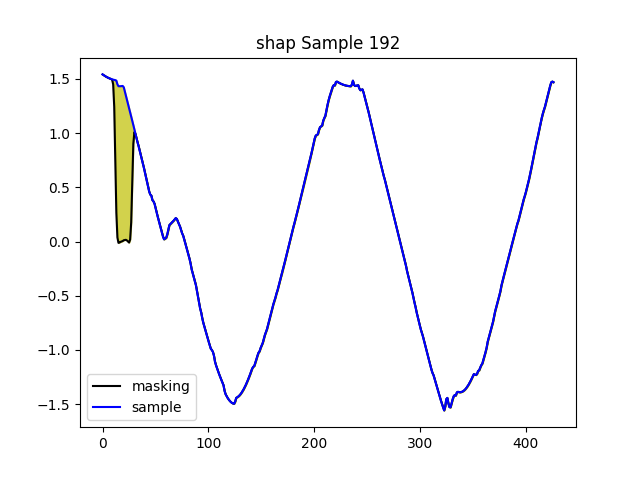} &
        \includegraphics[width=0.32\textwidth]{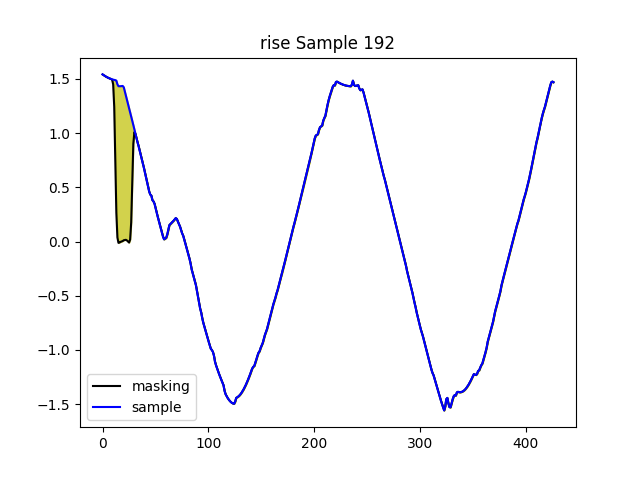} \\
        (e) LIME & (f) SHAP & (g) RISE
    \end{tabular}
    \caption{Explanation samples for Yoga dataset, Class 1, Sample 192}
    \label{fig:yoga_samples_192}
\end{figure*}

\clearpage
\subsection{Ford A Dataset}
\begin{figure*}[!ht]
    \centering
    \begin{tabular}{@{}ccc@{}}
        \multicolumn{3}{c}{\includegraphics[width=0.48\textwidth]{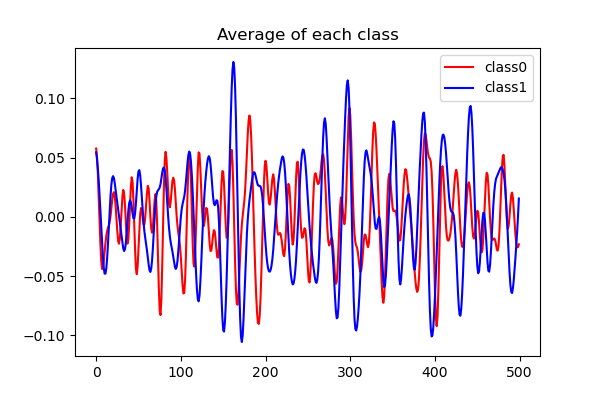}} \\
        \multicolumn{3}{c}{(a) Average of test samples in each class} \\[12pt]
        \includegraphics[width=0.32\textwidth]{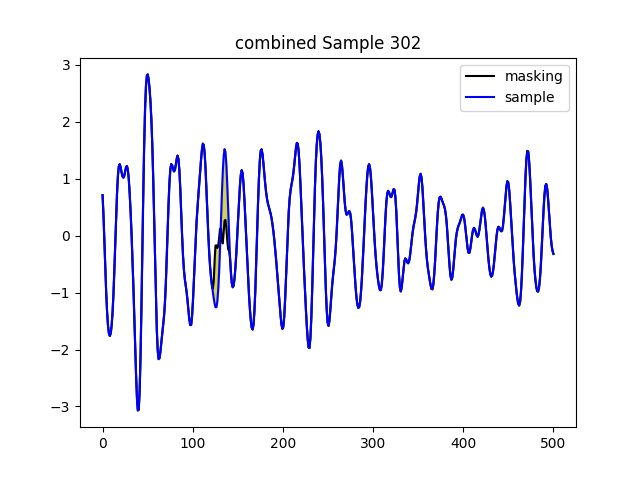} &
        \includegraphics[width=0.32\textwidth]{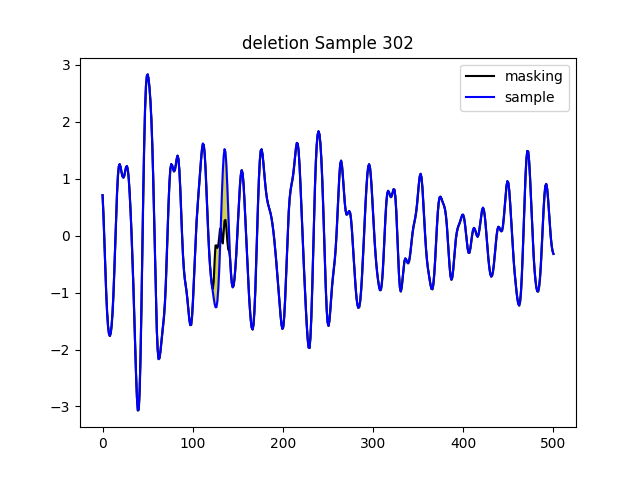} &
        \includegraphics[width=0.32\textwidth]{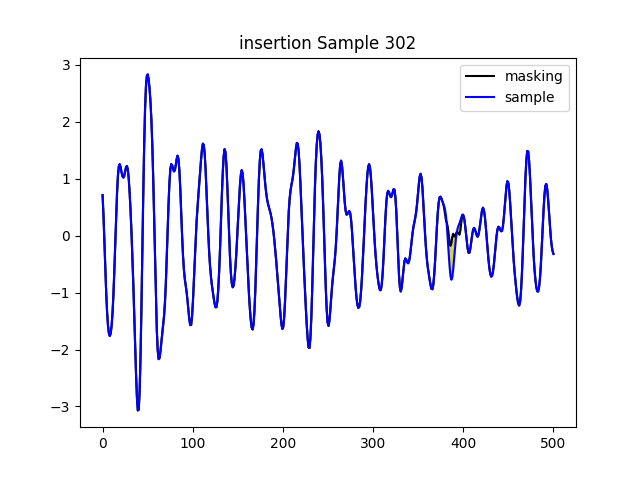} \\
        (b) Combined & (c) Deletion & (d) Insertion \\[12pt]
        \includegraphics[width=0.32\textwidth]{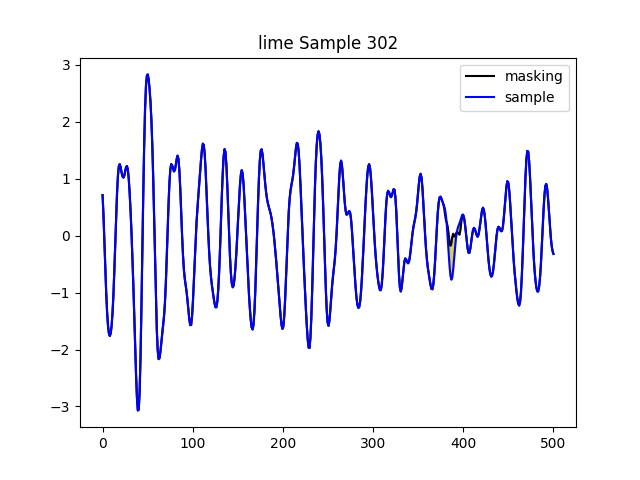} &
        \includegraphics[width=0.32\textwidth]{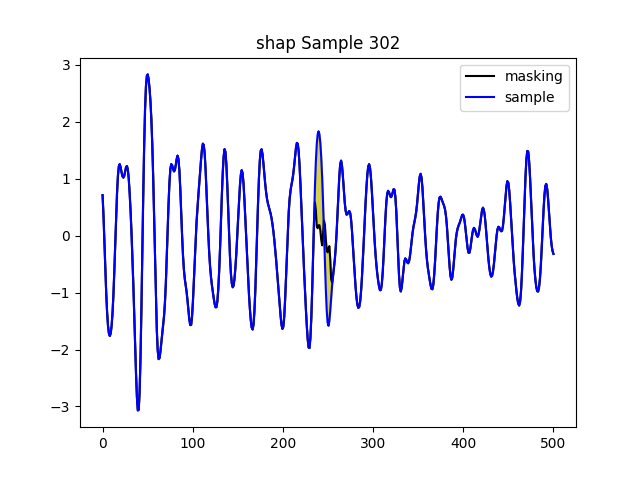} &
        \includegraphics[width=0.32\textwidth]{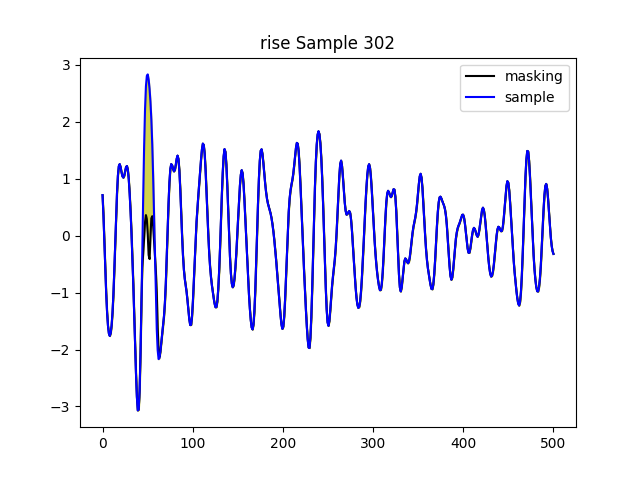} \\
        (e) LIME & (f) SHAP & (g) RISE
    \end{tabular}
    \caption{Explanation samples for Ford A dataset, Class 1, Sample 302}
    \label{fig:forda_samples}
\end{figure*}

\onecolumn 
\section{Classification Results}
\label{apd:first}

Here, we present the average and standard deviations of Accuracy, F1, Precision, and Recall for all three models in each of our dataset. We also visualize the confusion matrices of the two lowest performing models: Bi-LSTM model in CincECGTorso dataset, and Transformer model in MixedShapes dataset.

\begin{table}[h]
  \caption{Average Precision, Recall, F1, and Accuracy for all datasets}\label{tab:classification_results}
  \setlength{\tabcolsep}{4pt}
  \small
  \begin{tabular}{lcccc}
  \toprule
    & \multicolumn{4}{c}{Classification Metrics} \\
    \cmidrule(lr){2-5}
    & Precision & Recall & F1 & Accuracy  \\
  \midrule
  Synthetic Dataset & $1.0$ & $1.0$ & $1.0$ & $1.0$ \\
  CincECGTorso & $0.980 \pm 0.03$ & $0.979 \pm 0.05$ & $0.978 \pm 0.03$ & $0.977 \pm 0.03$ \\
  TwoPatterns & $0.984 \pm 0.03$ & $0.986 \pm 0.03$ & $0.985 \pm 0.02$ & $0.987 \pm 0.02$ \\
  MixedShapes & $0.945 \pm 0.04$ & $0.942 \pm 0.06$ & $0.943 \pm 0.05$ & $0.95 \pm 0.02$ \\
  Arrowhead & $1.0$ & $1.0$ & $1.0$ & $1.0$ \\
  Strawberry & $1.0$ & $1.0$ & $1.0$ & $1.0$ \\
  Yoga & $1.0$ & $1.0$ & $1.0$ & $1.0$  \\
  Ford A & $1.0$ & $1.0$ & $1.0$ & $1.0$  \\
  Ford B & $1.0$ & $1.0$ & $1.0$ & $1.0$  \\
  GunpointMaleFemale & $1.0$ & $1.0$ & $1.0$ & $1.0$  \\
  \bottomrule
  \end{tabular}
\end{table}

As shown in Table \ref{tab:classification_results}, our bi-LSTM, CNN, and Transformer classifiers performed perfectly well on the test set for the Synthetic, Arrowhead, Strawberry, Yoga, Ford A, and Ford B datasets. For the CincECGTorso, TwoPatterns, and MixedShapes, the averages of all three models were above $0.94$ for all classification metrics, which is still high enough to apply SpectralX framework and compare explanation quality of FIA and various perturbation methods.

\begin{figure}[h]
  \centering
  \includegraphics[width=0.48\textwidth]{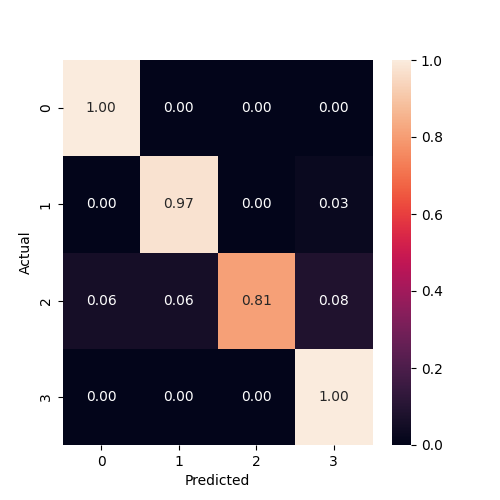}
  \includegraphics[width=0.48\textwidth]{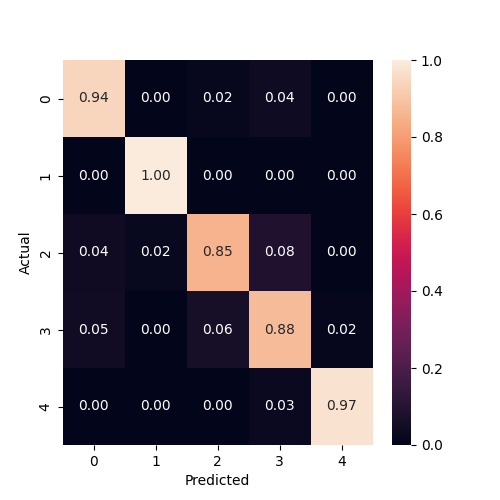}
  \caption{Confusion Matrices. Left: CincECGTorso Bi-LSTM model. Right: MixedShapes Transformer model.}
  \label{classification_figure1}
\end{figure}

The Confusion Matrices shown in Figure \ref{classification_figure1} visualize the comparison between actual and predicted classifications for the two most under-performing classification models. For the Bi-LSTM classification model in the CincECGTorso dataset, the model has the most difficulty classifying class two predictions. For the Transformer classification model, classes two and three predictions are under-performing compared to classes 0, 1, and 4. 

\clearpage
\section{Faithfulness and Robustness Details}
\label{apd:one_half}
In Table \ref{tab:faithfulness_total}, the average robustness values of Time-Frequency domain tend to be lower than the Time-domain for many perturbation methods: KernelSHAP, RISE, and Combined method. We hypothesize that this is because the Arrowhead datasets contain constant frequency values over the entire duration of each samples, and therefore the performance is better in the time-domain. However, except for the Arrowhead dataset, Faithfulness values are better in the Time-Frequency domain for the Combined method. 
In Table \ref{tab:robustness_total}, the robustness values of time-frequency domain explanations are greater than time domain explanations for all datasets and methods. We hypothesize this is due to the greater effect given by the noise in raw signals compared to the noise in STFT output magnitudes. 

{\small
\begin{table*}[ht]
  {\caption{Average Faithfulness for each dataset in Time domain (T) and Time-Frequency (TF) domain. Values indicate averages of three different classifiers.}\label{tab:faithfulness_total}
  \setlength{\tabcolsep}{10pt}
  \begin{tabular}{lcccccccccc}
  \toprule
    \multicolumn{2}{c}{\multirow{2}{*}{}} &
    \multicolumn{9}{c}{Average Faithfulness}
    \\\cmidrule(lr){3-11}
      & & CincECG. & Two. & Mixed. & Arrowhead & Strawberry & Yoga & Ford A & Ford B & Gun
  \\\hline
  \midrule
  \\\hline
  \multirow{2}{*}{LIME} & TF. & $0.043$ & $0.071$ & $0.041$ & $0.183$ & $0.146$ & $0.198$ & $0.218$ & $0.188$ & $0.228$
  \\
  & T. & $0.046$ & $0.048$ & $0.043$ & $0.166$ & $0.139$ & $0.192$ & $0.212$ & $0.182$ & $0.222$
  \\\hline
  \multirow{2}{*}{KernelSHAP} & TF. & $0.033$ & $0.044$ & $0.047$ & $0.159$ & $0.128$ & $0.175$ & $0.195$ & $0.165$ & $0.205$
  \\
  & T. & $0.028$ & $0.060$ & $0.045$ & $0.162$ & $0.131$ & $0.178$ & $0.198$ & $0.168$ & $0.208$
  \\\hline
  \multirow{2}{*}{RISE} & TF. & $0.041$ & $0.042$ & $0.042$ & $0.155$ & $0.143$ & $0.206$ & $0.226$ & $0.196$ & $0.236$ 
  \\
  & T. & $0.037$ & $0.039 $ & $0.039$ & $0.161$ & $0.135$ & $0.191$ & $0.211$ & $0.181$ & $0.221$
  \\\hline
  \multirow{2}{*}{LIMESegment} & TF. & - & - & - & - & -   
  \\
  & T. & $0.046$ & $0.065$ & $0.054$ & $0.163$ & $0.148$ & $0.204$ & $0.224$ & $0.194$ & $0.236$
  \\\hline
  \\\hline
  \multirow{2}{*}{Insertion} & TF. & $0.037$ & $0.032$ & $0.046$ & $0.149$ & $0.115$ & $0.154$ & $0.174$ & $0.144$ & $0.184$
  \\
  & T. & $0.035$ & $0.035$ & $0.052 $ & $0.147 $ & $0.107$ & $0.137$ & $0.157$ & $0.127$ & $0.167$ 
  \\\hline
  \multirow{2}{*}{Deletion} & TF. & $0.047$ & $0.083$ & $0.061$ & $0.177$ & $0.149$ & $0.196$ & $0.216$ & $0.186$ & $0.226$
  \\
  & T. & $0.046$ & $0.064$ & $0.055$ & $0.174$ & $0.148$ & $0.201$ & $0.221$ & $0.191$ & $0.231$
  \\\hline
  \multirow{2}{*}{Combined} & TF. & $0.042$ & $0.088$ & $0.071$ & $0.189$ & $0.154$ & $0.203$ & $0.218$ & $0.195$ & $0.226$
  \\
  & T. & $0.041$ & $0.077$ & $0.068$ & $0.192$ & $ 0.147$ & $0.190$ & $0.210$ & $0.180$ & $0.220$
  \\\hline
  \bottomrule
  \end{tabular}
}
\end{table*}

{\small
\begin{table*}[ht]
  {\caption{Average Robustness for each dataset in Time (T) and Time-Frequency (TF) domain. Values indicate averages of three different classifiers.}\label{tab:robustness_total}
  \setlength{\tabcolsep}{10pt}
  \begin{tabular}{lcccccccccc}
  \toprule
    \multicolumn{2}{c}{\multirow{2}{*}{}} &
    \multicolumn{9}{c}{Average Robustness}
    \\\cmidrule(lr){3-11}
      & & CincECG. & Two. & Mixed. & Arrowhead & Strawberry & Yoga & Ford A & Ford B & Gun
  \\\hline
  \midrule
  \\\hline
  \multirow{2}{*}{LIME} & TF. & $0.724$ & $0.662$ & $0.827$ & $0.684$ & $0.650$ & $0.498$ & $0.909$ & $0.766$ & $0.489$
  \\
  & T. & $0.613$ & $0.553$ & $0.716$ & $0.574$ & $0.541$ & $0.391$ & $0.796$ & $0.655$ & $0.382$
  \\\hline
  \multirow{2}{*}{KernelSHAP} & TF. & $0.732$ & $0.674$ & $0.830$ & $0.694$ & $0.662$ & $0.519$ & $0.907$ & $0.772$ & $0.510$
  \\
  & T. & $0.626$ & $0.560$ & $0.737$ & $0.584$ & $0.548$ & $0.385$ & $0.824$ & $0.671$ & $0.375$
  \\\hline
  \multirow{2}{*}{RISE} & TF. & $0.717$ & $0.650$ & $0.831$ & $0.674$ & $0.636$ & $0.470$ & $0.919$ & $0.763$ & $0.460$ 
  \\
  & T. & $0.591$ & $0.535 $ & $0.684$ & $0.555$ & $0.524$ & $0.387$ & $0.758$ & $0.629$ & $0.378$
  \\\hline
  \multirow{2}{*}{LIMESegment} & TF. & - & - & - & - & -   
  \\
  & T. & $0.639$ & $0.568$ & $0.760$ & $0.593$ & $0.554$ & $0.377$ & $0.855$ & $0.689$ & $0.366$
  \\\hline
  \\\hline
  \multirow{2}{*}{Insertion} & TF. & $0.733$ & $0.673$ & $0.836$ & $0.694$ & $0.661$ & $0.511$ & $0.916$ & $0.775$ & $0.502$
  \\
  & T. & $0.605$ & $0.523$ & $0.745 $ & $0.552 $ & $0.507$ & $0.303$ & $0.854$ & $0.662$ & $0.290$ 
  \\\hline
  \multirow{2}{*}{Deletion} & TF. & $0.729$ & $0.658$ & $0.844$ & $0.683$ & $0.644$ & $0.467$ & $0.934$ & $0.779$ & $0.456$
  \\
  & T. & $0.622$ & $0.546$ & $0.749$ & $0.573$ & $0.531$ & $0.344$ & $0.850$ & $0.674$ & $0.332$
  \\\hline
  \multirow{2}{*}{Combined} & TF. & $0.750$ & $0.696$ & $0.841$ & $0.715$ & $0.685$ & $0.552$ & $0.912$ & $0.787$ & $0.544$
  \\
  & T. & $0.601$ & $0.545$ & $0.694$ & $0.565$ & $ 0.534$ & $0.397$ & $0.768$ & $0.639$ & $0.388$
  \\\hline
  \bottomrule
  \end{tabular}
}
\end{table*}

\clearpage



\end{document}